\documentclass[11pt]{article}

\usepackage[preprint]{acl}

\usepackage{times}
\usepackage{latexsym}
\usepackage{amsfonts}

\usepackage[T1]{fontenc}
\usepackage[utf8]{inputenc}

\usepackage{microtype}

\usepackage{inconsolata}

\usepackage{graphicx}

\usepackage{enumitem}
\usepackage{colortbl}
\usepackage{amsmath}
\usepackage{booktabs}
\usepackage{multirow}
\usepackage{microtype}
\usepackage{graphicx}
\usepackage{subcaption}

\usepackage{xcolor}
\usepackage{listings}
\usepackage[most]{tcolorbox}

\usepackage{mathtools}
\usepackage{amsthm}
\usepackage{array}       % 表格列格式增强（推荐）
\usepackage[table]{xcolor}
\usepackage{newunicodechar}
\usepackage{pifont}

\definecolor{bgblue}{RGB}{235, 245, 250}   % Light Blue for Basic
\definecolor{bggreen}{RGB}{235, 250, 235}  % Light Green for Sparse/Matrix
\definecolor{bgred}{RGB}{250, 235, 235}    % Light Red for Data-Driven
\definecolor{top1bg}{RGB}{255, 200, 110}   % Top-1: saturated amber/gold
\definecolor{top2bg}{RGB}{255, 225, 170}   % Top-2: medium amber
\definecolor{top3bg}{RGB}{255, 242, 215}   % Top-3: pale amber

\newcommand{\cmark}{\ding{51}} % Checkmark
\newcommand{\xmark}{\ding{55}} % Cross

\title{Exploring Information Seeking Agent Consolidation}

\author{
 \textbf{Guochen Yan\textsuperscript{1,*}},
 \textbf{Jialong Wu\textsuperscript{1,*}},
 \textbf{Zhengwei Tao\textsuperscript{1}},
 \textbf{Bo Li\textsuperscript{1}},
 \textbf{Qintong Zhang\textsuperscript{1}},
 \textbf{Jiahao Xu\textsuperscript{2}},
 \\
 \textbf{Haitao Mi\textsuperscript{2}},
 \textbf{Yuejian Fang \textsuperscript{1}},
 \textbf{Qingni Shen\textsuperscript{1}},
 \textbf{Wentao Zhang\textsuperscript{1}},
 \textbf{Zhonghai Wu\textsuperscript{1}},
\\
\\
 \textsuperscript{1}Peking University,
 \textsuperscript{2}Tencent
\\
}

\begin{document}
\maketitle
\begin{abstract}
Information-seeking agents have emerged as a powerful paradigm for knowledge-intensive tasks, yet today's systems remain specialized for the open web, documents, or local knowledge bases, hindering scalable and cross-domain deployment. We present the first systematic empirical study of consolidating these information-seeking agents into a single foundation agentic model. We compare two paradigms---\emph{data-level mixing}, which trains a unified model on a mixture of datasets, and \emph{parameter-level merging}, which merges independently trained experts in parameter space---across 3 training scenarios, evaluating \textbf{26} representative parameter-level methods on \textbf{10} benchmarks. To compare across heterogeneous benchmarks, we introduce a geometric Composite Score and an Imbalance Score that describe overall performance and task skew. Our analysis shows that (i) well-designed parameter-level merging attains parity with data mixing at a fraction of its training cost and is order-agnostic; (ii) parameter-level merging structurally preserves out-of-domain capabilities that data mixing universally forgets; and (iii) cross-scenario stability is strongly tied to consolidation quality. We distil our observations into a method-selection guide and design principles for next-generation merging operators.
\end{abstract}

\section{Introduction}

\begin{figure*}[t]
  \centering
   \includegraphics[width=\linewidth]{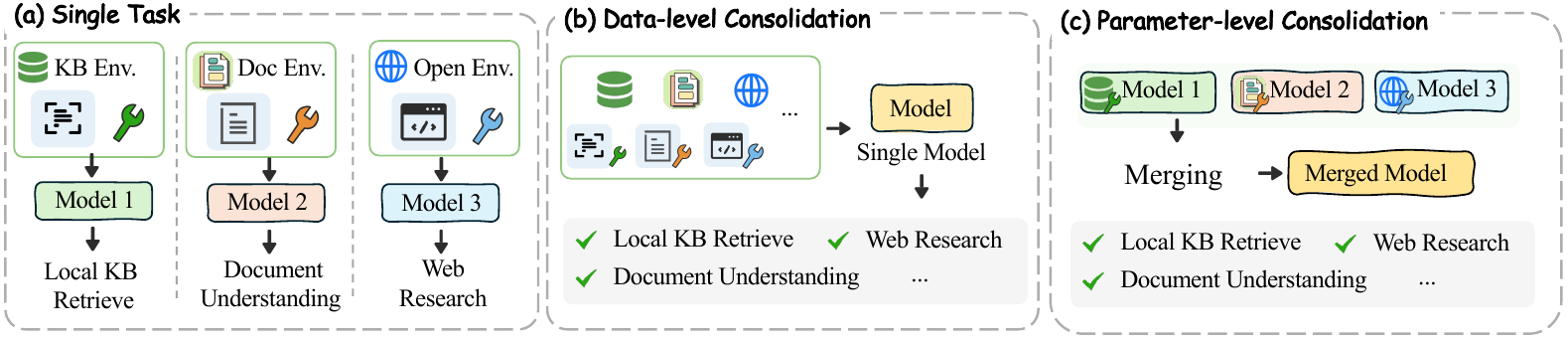}
   \caption{Comparison of three information-seeking agent consolidation paradigms.
\textbf{(a) Single-task training}, where separate agents are independently trained for local knowledge-base retrieval $\mathcal{D}_{\text{rag}}$, document understanding $\mathcal{D}_{\text{doc}}$, and open-web search $\mathcal{D}_{\text{web}}$, in their respective environments.
\textbf{(b) Data-level consolidation}, which unifies heterogeneous agent trajectories into a single training set $\mathcal{D}_{\text{all}}$ and learns a single model via joint multi-task training.
\textbf{(c) Parameter-level consolidation}, which first trains environment-specific expert models as in (a) and then merges them in parameter space to obtain a unified agent without joint retraining.}
   \label{fig:overall}
\end{figure*}

Information-seeking agents that interleave retrieval, reasoning, and tool use have become the dominant paradigm for knowledge-intensive tasks. The literature has converged on three specialised classes, each tied to its native environment: \emph{Web agents} interacting with search engines and live pages~\cite{wu2025webdancer,li2025websailor,websailor-v2}; \emph{Document agents} reasoning over multi-modal documents~\cite{zhang2026docdancer,sun2025docagent,zhu2025doclens}; and \emph{RAG agents} operating over local knowledge bases~\cite{jin2025search,tao2026ragshaper}. Real-world deployments increasingly demand all three capabilities in a single model, since maintaining a separate expert per domain forces ad hoc routing, multiplies serving costs, and prevents capabilities from compounding. Consolidating the three experts into a unified \emph{foundation agentic model} (Figure~\ref{fig:overall}) is therefore both timely and underexplored.

Agent consolidation is more challenging than the model-merging settings studied in prior work~\cite{yang2026model,yadav2024matters}: a consolidated agent must interact with heterogeneous tool APIs, preserve multi-step planning across drastically different action distributions, and avoid forgetting any individual expert. Two paradigms currently compete. \emph{Data-level consolidation} jointly trains on a mixture of all task corpora. \emph{Parameter-level consolidation} merges independently trained experts in weight space~\cite{wang2025ui,team2025introducing}. The former is empirically strong but data- and compute-hungry, while the agentic effectiveness, robustness, and trade-offs of the latter remain poorly characterised.

To bridge the gap, we present the first systematic empirical study of information-seeking agent consolidation, organised around four research questions. \textbf{(RQ1) Performance}: can parameter-level merging match data-level mixing's performance? \textbf{(RQ2) Generalization}: does either paradigm retain out-of-domain capability? \textbf{(RQ3) Failure modes}: which scenarios break which methods? \textbf{(RQ4) Mechanisms}: which design factors govern merging effectiveness? We evaluate data-mixing baseline and \textbf{26} parameter-level methods, varying backbone scale and fine-tuning regime. For effective comparison, we introduce a geometric \emph{Composite Score} and an \emph{Imbalance Score} that describe overall quality and task imbalance, respectively.

Five headline findings emerge in our study. (F1) Well-designed parameter-level consolidation can attain parity with data mixing. (F2) Parameter-level merging better preserves general capabilities. (F3) Stability is structurally correlated to quality. (F4) Three design factors govern merging effectiveness: consolidation granularity, consensus-indicator informativeness, and expert task affinity. (F5) Parameter-level methods preserve the behavior of original expert agents. We distil these into a deployment guide and four design principles.

In summary, our contributions are as follows:
\begin{description}[
    style=standard,
    leftmargin=0pt,
    nosep
]
    \item[\textbullet\ ] \textbf{Unified information-seeking agent paradigm.} We formulate web search, document-grounded reasoning, and knowledge-base retrieval under a single paradigm, enabling a consistent abstraction across heterogeneous environments. (\S\ref{sec:unify},~\S\ref{sec:setup}).

    \item[\textbullet\ ] \textbf{Largest-scale systematic evaluation.} We benchmark \textbf{26} parameter-level consolidation methods against data mixing on 8 in-domain and 2 OOD benchmarks, with metrics ($G$, $\mathrm{Imb}$) that separate overall quality from imbalance (\S\ref{sec:setup}).

    \item[\textbullet\ ] \textbf{Novel empirical findings.} We identify a co-champion pair across 3 scenarios, OOD-retention advantage of parameter-level methods, stability-quality correlation, design factors governing merging effectiveness, and behavior analysis (\S\ref{sec:empirical}).

    \item[\textbullet\ ] \textbf{Design principles.} Based on our empirical study, we distil a constraint-driven method-selection guide and four design principles (\S\ref{sec:design_principles}).
    % open release?
\end{description}

\section{Related Works}
\noindent \textbf{Information Seeking Agents.} RAG has emerged as the primary paradigm for grounding LLMs in external knowledge bases~\cite{wang2024searching}, with recent advances in multi-hop reasoning~\cite{li2024structrag} and adaptive memory-based optimization~\cite{qin2025towards}. Document-centric agents tackle visually-rich documents such as PDFs and slides through vision-language models~\cite{faysse2024colpali} and multi-modal retrieval frameworks~\cite{tanaka2025vdocrag,han2025mdocagent}, enabling precise extraction from complex layouts and multi-page contexts~\cite{ma2024mmlongbench,jin2025slideagent}. Web agents navigate dynamic websites to gather open-domain information~\cite{deng2023mind2web,wu2025webdancer}, relying on planning and rollback mechanisms for robustness in open-web environments~\cite{wei2025browsecomp,zhang2025enhancing}.

\noindent \textbf{Model Consolidation.} Specialized capabilities can be consolidated into unified systems through two paradigms: \emph{data-level mixture} and \emph{parameter-level merging}. Data-level methods build on multi-task instruction tuning, where training on massive task mixtures yields broad generalization~\cite{chung2024scaling,longpre2023flan}. Recent work adapts this paradigm to agentic settings by scaling across heterogeneous environments~\cite{fang2025towards} and applying reinforcement learning~\cite{xi2025agentgym}. Parameter-level consolidation, in parallel, builds on classical parameter merging~\cite{shoemake1985animating,utans1996weight}, originally used to combine distinct NLP abilities~\cite{yu2024language,wan2025fusechat,yang2026model} into foundation models without the computational cost of joint retraining. Recent work extends this paradigm to agentic settings~\cite{liao2025marft,maiti2025souper}. Our work systematically evaluates both strategies to overcome the fragmentation of existing information-seeking agents and enable cross-domain scalability.

\section{Preliminaries and Setup}
\label{sec:unify}
\subsection{Task Definition}

We study information-seeking agents that answer a natural-language query by retrieving and reasoning over external knowledge sources. Formally, let $q \in \mathcal{Q}$ denote a user query and $y \in \mathcal{Y}$ the target answer; the agent is a parameterized policy $\pi_\theta$ that induces a mapping
\begin{equation}
    \pi_\theta : \mathcal{Q} \times \mathcal{K} \rightarrow \mathcal{Y},
\end{equation}
where $\mathcal{K}$ denotes the accessible knowledge environment and $\theta$ the learnable parameters. We categorize information-seeking tasks into three classes by the structure of $\mathcal{K}$ and the corresponding \emph{retrieval interfaces}, which together define the agent's action space $\mathcal{A}$. Across all settings the agent follows the same \texttt{ReAct} paradigm~\citep{yao2022react}, differing only in how knowledge is accessed and observed; the resulting $T$-iteration interaction history is
\begin{equation}
\mathcal{H}_T = (\tau_0, a_0, o_0, \dots, \tau_i, a_i, o_i, \dots, \tau_T, a_T),
\end{equation}
where at each step $t$ the agent generates an internal thought $\tau_t$, selects an action $a_t \in \mathcal{A}$, and the environment returns an observation $o_t$.

\noindent \textbf{Online Open-ended Web Search.} The agent interacts with a dynamic, partially observable environment $\mathcal{E}_{\text{web}}$ via two actions: \emph{search}, parameterized by a query and returning a list of titles and snippets, and \emph{visit}, parameterized by a goal and a URL and returning evidence extracted from the corresponding webpage. Web agents reduce problem uncertainty by leveraging information from the web, which requires strong problem decomposition and associative reasoning.

\noindent \textbf{Document-Grounded Understanding.} The agent is provided with a document that may contain multimodal elements such as text, tables, figures, or images, and alternates between \emph{Search} actions, which provide global textual signals over the document collection, and \emph{Read} actions, which perform fine-grained, localized extraction from selected sections. Doc agents therefore focus on localized information extraction, long-context understanding, and coherent reasoning within a single source, rather than retrieval or external exploration.

\noindent \textbf{Local Knowledge Base Retrieval and Generation.} The agent operates over a fixed and curated corpus $\mathcal{K}_{\text{rag}} = \{d_1, d_2, \dots, d_N\}$ whose documents are static and known a priori (e.g., Wikipedia); at each time step it accesses $\mathcal{K}_{\text{rag}}$ through a dense retrieval interface implemented as an embedding index. RAG agents couple retrieval with generation, emphasizing faithful grounding and synthesis of retrieved content.

\begin{table}[t]
    \small
    \centering
    \caption{Comparison of three different information-seeking agents.}
    \label{tab:comparison}
    \begin{tabular}{ccc}
    \toprule
      Agent Model $\pi_\theta$ & Environment $\mathcal{K}$ & Tool Set $\mathcal{A}$ \\
      \midrule
       Web  & Internet & \textit{Search}, \textit{Visit}\\
       Doc  & Document & \textit{Search}, \textit{Read} \\
       RAG  & Wikipedia & \textit{Dense Retrieval} \\
    \bottomrule
    \end{tabular}
\end{table}
Table~\ref{tab:comparison} summarizes these distinctions; per-agent implementation details are deferred to Appendix~\ref{app:agents}. Prior work typically trains an agent separately on each dataset, producing parameters $\theta_{\text{web}}$, $\theta_{\text{doc}}$, and $\theta_{\text{rag}}$ on $\mathcal{D}_{\text{web}}$, $\mathcal{D}_{\text{doc}}$, and $\mathcal{D}_{\text{rag}}$ respectively. In this work, we investigate how these three information-seeking agents can be consolidated into a unified framework (Figure~\ref{fig:overall}). Existing approaches to consolidation operate at either the data level or the parameter level.

\noindent \textbf{Data-level Consolidation.} Data-level consolidation operates directly on the training data: we merge the ReAct-style trajectory datasets collected from the three environments into a single consolidated dataset
\begin{equation}
\mathcal{D}_{\text{all}} = \mathcal{D}_{\text{web}} \cup \mathcal{D}_{\text{doc}} \cup \mathcal{D}_{\text{rag}},
\end{equation}
and then fine-tune a unified agent on $\mathcal{D}_{\text{all}}$, enabling it to handle multiple information environments within a single set of parameters.

\noindent \textbf{Parameter-level Consolidation.} Parameter-level consolidation instead merges multiple domain-specialized models in the parameter space: we first train a set of models independently, each specialized for one environment, and then consolidate them by merging their parameters into a generalized model
\[
\theta^{(\mathrm{merge})}
=
\mathrm{merge}\!\left(
\theta_{\text{web}},
\theta_{\text{doc}},
\theta_{\text{rag}}
\right),
\]
where $\mathrm{merge}(\cdot)$ denotes a generic parameter merging operator. One straightforward instantiation is \emph{linear average}~\cite{wortsman2022model}, merging parameters as a convex combination
\begin{equation}
\begin{aligned}
\theta^{(\mathrm{merge})}
&=
\alpha_{\text{web}} \theta_{\text{web}}
+
\alpha_{\text{doc}} \theta_{\text{doc}}
+
\alpha_{\text{rag}} \theta_{\text{rag}}
 \\
\text{s.t.}\quad
&\alpha_{\text{web}} + \alpha_{\text{doc}} + \alpha_{\text{rag}} = 1, \quad \alpha_{\cdot} \ge 0 .
\end{aligned}
\end{equation}
More advanced merging operators that go beyond convex combinations are presented in \S\ref{sec:setup}.

\begin{figure}[t]
  \centering
   \includegraphics[width=0.9\linewidth]{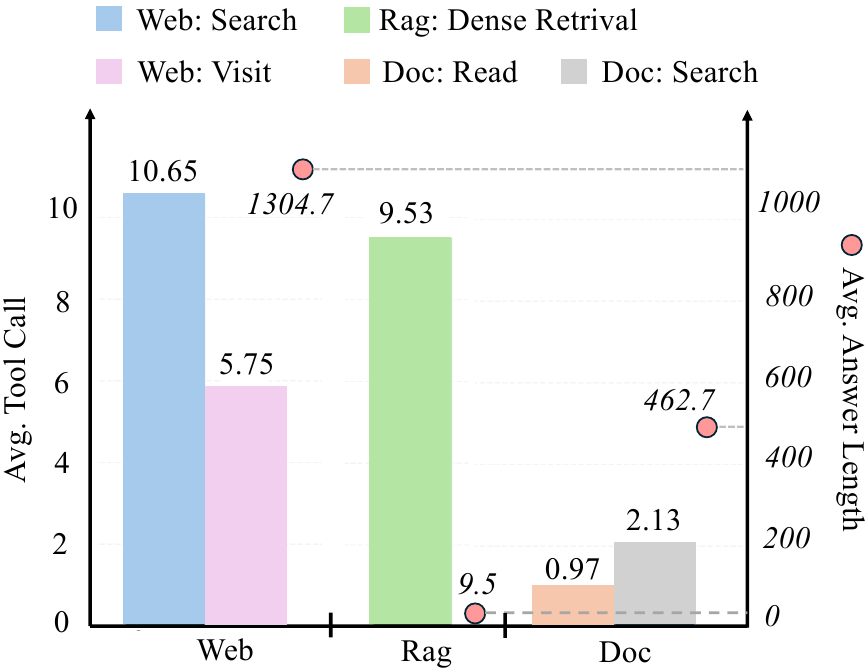}
   \caption{Average tool usage frequency and average answer length across different information-seeking settings in training data.
   }
   \label{fig:tool_answer}
   \vspace{-5mm}
\end{figure}

\begin{table}[t]
\centering
\caption{Taxonomy and comparison of parameter-level consolidation methods. We highlight three categories: \colorbox{bgblue}{Basic Interpolation}, \colorbox{bggreen}{Interference Resolution}, and \colorbox{bgred}{Data-Driven Optimization}. Consensus Strategy specifies the mechanism employed to mitigate interference and harmonize conflicting parameters.}
\label{tab:tax_merge_methods}
\renewcommand{\arraystretch}{1.2} % Increase row height for readability
\setlength{\tabcolsep}{2pt}
\resizebox{\linewidth}{!}{
\begin{tabular}{lcccc}
\toprule
\textbf{Method} & \textbf{Sparsification} & \textbf{Granularity} & \textbf{Consensus Strategy} & \textbf{Data-Free} \\ 
\midrule

% --- Group 1: Basic Interpolation
\rowcolor{bgblue}
\textbf{Average} & -- & Model & -- & \textcolor{green!60!black}{\cmark} \\
\rowcolor{bgblue}
\textbf{SLERP} & -- & Model & -- & \textcolor{green!60!black}{\cmark} \\
\rowcolor{bgblue}
\textbf{MetaGPT} & -- & Model & -- & \textcolor{green!60!black}{\cmark} \\
\rowcolor{bgblue}
\textbf{LiNeS} & -- & Layer & -- & \textcolor{green!60!black}{\cmark} \\

% --- Group 2: Interference Resolution (ordered consistently with Table~\ref{tab:performance_comparison})
\rowcolor{bggreen}
\textbf{DARE} & \cmark & Model & Random Dropout & \textcolor{green!60!black}{\cmark} \\
\rowcolor{bggreen}
\textbf{Breadcrumbs} & \cmark & Model & Random Dropout & \textcolor{green!60!black}{\cmark} \\
\rowcolor{bggreen}
\textbf{TIES} & \cmark & Matrix & Sign Consistency & \textcolor{green!60!black}{\cmark} \\
\rowcolor{bggreen}
\textbf{Consensus TA} & \cmark & Matrix & Target Alignment & \textcolor{green!60!black}{\cmark} \\
\rowcolor{bggreen}
\textbf{TADrop} & \cmark & Matrix & Adaptive Dropout & \textcolor{green!60!black}{\cmark} \\
\rowcolor{bggreen}
\textbf{CABS} & \cmark & Matrix & Disjoint Aggregation & \textcolor{green!60!black}{\cmark} \\
\rowcolor{bggreen}
\textbf{PCB Merging} & \cmark & Matrix & Sign Consistency & \textcolor{green!60!black}{\cmark} \\
\rowcolor{bggreen}
\textbf{DELLA} & \cmark & Matrix & Sign Consistency & \textcolor{green!60!black}{\cmark} \\
\rowcolor{bggreen}
\textbf{SCE} & \cmark & Matrix & Sign Consistency & \textcolor{green!60!black}{\cmark} \\
\rowcolor{bggreen}
\textbf{WIDEN} & -- & Matrix & Magnitude--Direction Disentangle & \textcolor{green!60!black}{\cmark} \\
\rowcolor{bggreen}
\textbf{RAM+} & -- & Layer & Permutation Alignment & \textcolor{green!60!black}{\cmark} \\
\rowcolor{bggreen}
\textbf{TSV} & -- & Matrix & Subspace Denoising & \textcolor{green!60!black}{\cmark} \\
\rowcolor{bggreen}
\textbf{ISO-CTS} & -- & Matrix & Spectral Isotropy & \textcolor{green!60!black}{\cmark} \\
\rowcolor{bggreen}
\textbf{IMPART} & \cmark & Matrix & Spectral Filtering & \textcolor{green!60!black}{\cmark} \\
\rowcolor{bggreen}
\textbf{WUDI} & -- & Matrix & Subspace Interference Min. & \textcolor{green!60!black}{\cmark} \\
\rowcolor{bggreen}
\textbf{DC-Merge} & -- & Matrix & Subspace Decomposition & \textcolor{green!60!black}{\cmark} \\
\rowcolor{bggreen}
\textbf{OrthoMerge-C} & -- & Matrix & Common-Subspace Orthogonalization & \textcolor{green!60!black}{\cmark} \\
\rowcolor{bggreen}
\textbf{OrthoMerge-G} & -- & Matrix & Global Orthogonalization & \textcolor{green!60!black}{\cmark} \\

% --- Group 3: Data-Driven Optimization
\rowcolor{bgred}
\textbf{AdaMerging} & -- & Layer & Entropy Minimization & \textcolor{red}{\xmark} \\
\rowcolor{bgred}
\textbf{RegMean++} & -- & Matrix & Activation Matching & \textcolor{red}{\xmark} \\
\rowcolor{bgred}
\textbf{CAT Merging} & -- & Matrix & Feature-Aware Trimming & \textcolor{red}{\xmark} \\
\rowcolor{bgred}
\textbf{SLTA} & -- & Matrix & Submodule Linearity & \textcolor{red}{\xmark} \\

\bottomrule
\end{tabular}
}
\end{table}

\section{Experimental Setup}
\label{sec:setup}
\noindent \textbf{Training Datasets.} Following~\cite{zhang2026docdancer,tao2026ragshaper,wu2025webdancer}, we construct training data for the three corresponding agents, collecting 4{,}500 ReAct-style agent trajectories per setting. As shown in Figure~\ref{fig:tool_answer}, the three settings exhibit substantially different tool-use and response profiles: Web has the highest tool-call frequency (reflecting iterative search and page visits), Document has the lowest (consistent with its localized search-and-read process), and RAG primarily relies on dense retrieval with fewer tool invocations and substantially shorter responses dictated by its short-form evaluation format. The information-seeking strategies, interaction dynamics, and response length vary significantly across settings. We train on Qwen3-30B-A3B-Think and Qwen3-4B-Think with 128$k$ context length. Implementation details are in Appendix~\ref{app:imple}.

\noindent \textbf{Benchmarks.}
For Web agent evaluation, we use \textbf{GAIA}~\cite{gaia} (on its 103-instance text-only subset, following~\cite{webthinker,wu2025webdancer}), together with BrowseComp (\textbf{BC})~\cite{wei2025browsecomp} and BrowseComp-zh (\textbf{BC-zh})~\cite{bc_zh}, randomly sampling 200 and 100 instances from BC and BC-zh, respectively, due to their high evaluation cost. All are scored by accuracy. For Doc agent evaluation, we use two multimodal long-context document question answering benchmarks, MMLongBenchDoc (\textbf{MMBD})~\cite{ma2024mmlongbench} and DocBench (\textbf{DocB})~\cite{zou2025docbench}, scored by LLM-as-Judge accuracy. For RAG agent evaluation we use \textbf{HotPotQA}~\cite{yang2018hotpotqa}, \textbf{AmbigQA}~\cite{min2020ambigqa}, and \textbf{Bamboogle}~\cite{press2023measuring}, reporting the standard Exact Match (EM) and F1 score. To additionally probe out-of-domain (OOD) capability retention beyond the three agentic tasks, we evaluate models on \textbf{AIME 2025}~\cite{aime2025} for competition mathematics and \textbf{LiveCodeBench} (\textbf{LCB}, \texttt{2408--2502})~\cite{jain2024livecodebench} for code generation, both reported using accuracy. Further benchmark details are in Appendix~\ref{sec:app_benchmark}.

\noindent \textbf{Aggregated Metrics.}
Since the benchmarks span heterogeneous scales and sizes, we aggregate them into two overall metrics in three stages: \textit{(i)} a sigmoid-squashed per-benchmark $z$-score, $s_{i,b}=1/(1+e^{-z_{i,b}})$ with $z_{i,b}=(x_{i,b}-\mu_b)/\sigma_b$, which removes scale differences and bounds pathological methods near $0$; \textit{(ii)} within-task arithmetic means $S_{i,\text{web}}, S_{i,\text{doc}}, S_{i,\text{rag}}$; \textit{(iii)} the \textbf{Composite Score} $G(i)=\sqrt[3]{S_{i,\text{web}}\,S_{i,\text{doc}}\,S_{i,\text{rag}}}$, the cross-task geometric mean, which (unlike an arithmetic mean) penalises any single-task collapse. To quantify across-task skew, we additionally report the \textbf{Imbalance Score} $\mathrm{Imb}(i)=\max\!\bigl(0,\,1-G(i)/\bar S_i\bigr)$ where $\bar S_i$ is the arithmetic mean of the three task scores. It equals $0$ for balanced profiles, tends to $1$ as performance concentrates on a single task, and is scale-invariant and outlier-robust by construction. All quantities are linearly mapped to $[0,100]$.

\noindent \textbf{Parameter-level Consolidation Methods.}
We evaluate \textbf{26 representative parameter-level consolidation methods}, organised into three categories in Table~\ref{tab:tax_merge_methods}: \textit{(i)}~\colorbox{bgblue}{Basic Interpolation} (Average, SLERP, MetaGPT, LiNeS), which linearly or spherically averages weights; \textit{(ii)}~\colorbox{bggreen}{Interference Resolution} (DARE, Breadcrumbs, TIES, Consensus TA, TADrop, CABS, PCB Merging, DELLA, SCE, WIDEN, RAM+, TSV, ISO-CTS, IMPART, WUDI, DC-Merge, OrthoMerge-C, OrthoMerge-G), which mitigates cross-task conflicts via sparsification, sign reconciliation, or subspace alignment; and \textit{(iii)}~\colorbox{bgred}{Data-Driven Optimization} (AdaMerging, RegMean++, CAT Merging, SLTA), which learns merging coefficients or weights from a few input data. Detailed descriptions are in Appendix~\ref{app:merge_methods}.

\begin{table*}[t]
    \small
    \centering
    \caption{Performance comparison using Qwen3-30B-A3B-Think as the backbone model. 
    The performance of Qwen3-4B-Think is reported in Table~\ref{tab:performance_comparison_4b}. \textbf{G} and \textbf{Imb.} are the Composite Score and Imbalance Score describe the overall performance and task imbalance on three tasks. For data- and parameter-level methods, the best, second-best, and third-best values are highlighted with \colorbox{top1bg}{darker}, \colorbox{top2bg}{medium}, and \colorbox{top3bg}{lighter} amber cells, respectively.
    }
    \label{tab:performance_comparison}
    \setlength{\tabcolsep}{3pt}
    \resizebox{\textwidth}{!}{%
    \begin{tabular}{l|ccc|cc|cccccc|cc|cc}
    \toprule
     \multirow{3}{*}{\textbf{Model}} & \multicolumn{3}{c|}{\textbf{Web}} & \multicolumn{2}{c|}{\textbf{Doc}} & \multicolumn{6}{c|}{\textbf{RAG}} & \multicolumn{2}{c|}{\textbf{Overall}} & \multicolumn{2}{c}{\textbf{OOD}}
     \\
     % 第二行表头：数据集名称
        & \textbf{GAIA} & \textbf{BC} & \textbf{BC-zh} & \textbf{MMBD} & \textbf{DocB} & \multicolumn{2}{c}{\textbf{HotPotQA}} & \multicolumn{2}{c}{\textbf{AmbigQA}} & \multicolumn{2}{c|}{\textbf{Bamboogle}} & \multirow{2}{*}{\textbf{G}$\uparrow$} & \multirow{2}{*}{\textbf{Imb.}$\downarrow$} & \textbf{AIME25} & \textbf{LCB} \\
     % 第三行表头：具体指标
        & \textit{Acc.} & \textit{Acc.} & \textit{Acc.} & \textit{Acc.} & \textit{Acc.} & \textit{EM} & \textit{F1} & \textit{EM} & \textit{F1} & \textit{EM} & \textit{F1} & & & \textit{Acc.} & \textit{Acc.} \\
    \midrule
    \multicolumn{16}{l}{\textit{\textbf{Expert Agent}}} \\
    \arrayrulecolor{black!20}\midrule
    Web Agent ($\mathcal{D}_{\text{web}}$) & 62.14 & 21.00 & 24.00 & 55.82 & 79.58 & 34.10 & 50.09 & 44.40 & 60.87 & 44.00 & 61.60 & 62.96 & 0.56 & 81.24 & 54.10 \\
     Doc Agent ($\mathcal{D}_{\text{doc}}$) & 68.93 & 13.50 & 27.00 & 64.18 & 81.79 & 28.30 & 46.21 & 28.90 & 50.80 & 41.60 & 60.98 & 61.73 & 0.03 & 85.00 & 60.49 \\
     RAG Agent ($\mathcal{D}_{\text{rag}}$) & 23.30 & 3.50 & 11.20 & 30.03 & 46.01 & 48.00 & 60.98 & 62.60 & 73.74 & 59.20 & 71.00 & 30.77 & 25.32 & 83.34 & 59.57 \\
    \arrayrulecolor{black}\midrule
    \multicolumn{16}{l}{\textit{\textbf{Data-level Consolidation}}} \\
    \arrayrulecolor{black!20}\midrule
     Data Mixing ($\mathcal{D}_{\text{all}}$) & \cellcolor{top1bg}64.08 & \cellcolor{top1bg}28.00 & \cellcolor{top1bg}34.00 & 63.59 & \cellcolor{top1bg}83.29 & 38.00 & 42.53 & 49.50 & 58.84 & \cellcolor{top1bg}53.10 & 60.20 & \cellcolor{top1bg}71.06 & 0.37 & 80.21 & 55.32 \\
    \arrayrulecolor{black}\midrule
    \multicolumn{16}{l}{\textit{\textbf{Parameter-level Consolidation}}} \\
    \arrayrulecolor{black!20}\midrule
     \cellcolor{bgblue}Average & 60.19 & 18.50 & 29.00 & 63.40 & 81.40 & 18.60 & 28.93 & 26.60 & 38.23 & 28.00 & 38.82 & 53.03 & 3.29 & 83.75 & \cellcolor{top3bg}61.40 \\
     \cellcolor{bgblue}SLERP & 60.19 & 18.00 & 25.00 & 62.66 & \cellcolor{top3bg}81.67 & 14.80 & 26.52 & 21.60 & 34.79 & 18.40 & 32.96 & 47.63 & 5.99 & 82.51 & 57.75 \\
     \cellcolor{bgblue}MetaGPT & 56.31 & \cellcolor{top3bg}22.00 & \cellcolor{top2bg}33.00 & 63.40 & \cellcolor{top2bg}81.85 & 13.20 & 25.65 & 18.90 & 33.55 & 19.20 & 32.19 & 48.91 & 7.96 & 84.38 & \cellcolor{top2bg}61.70 \\
     \cellcolor{bgblue}LiNeS & 58.25 & 18.00 & \cellcolor{top2bg}33.00 & 60.35 & 80.22 & 22.90 & 36.20 & 25.50 & 41.63 & 20.00 & 35.83 & 53.12 & 2.86 & 81.26 & 55.32 \\
     \cellcolor{bggreen}DARE & \cellcolor{top3bg}62.14 & 19.50 & 24.00 & 63.31 & 80.58 & 18.70 & 29.56 & 25.00 & 37.66 & 23.20 & 37.83 & 51.21 & 3.63 & \cellcolor{top3bg}84.60 & 58.05 \\
     \cellcolor{bggreen}Breadcrumbs & 55.34 & 10.00 & 19.00 & 58.68 & 77.77 & 26.10 & 39.14 & 28.10 & 42.25 & 23.20 & 41.29 & 47.86 & 0.62 & 78.34 & 44.68 \\
     \cellcolor{bggreen}TIES & \cellcolor{top3bg}62.14 & 14.50 & 24.00 & 60.26 & 79.76 & 25.00 & 38.48 & 29.60 & 44.53 & 32.00 & 47.22 & 54.62 & 0.42 & 80.01 & 50.46 \\
     \cellcolor{bggreen}Consensus TA & 13.59 & 0.50 & 6.00 & 24.49 & 31.58 & 13.50 & 19.77 & 19.60 & 30.21 & 20.00 & 28.94 & 14.64 & 7.05 & 54.16 & 24.01 \\
     \cellcolor{bggreen}TADrop & 54.37 & 12.00 & 18.00 & 59.61 & 77.77 & 23.50 & 37.12 & 25.60 & 39.70 & 22.40 & 37.84 & 46.74 & 1.16 & 79.58 & 50.15 \\
     \cellcolor{bggreen}CABS & \cellcolor{top1bg}64.08 & 19.50 & 29.00 & 61.92 & 80.76 & 24.10 & 42.89 & 24.20 & 46.45 & 32.80 & 55.95 & 59.50 & 0.76 & 79.59 & 51.98 \\
     \cellcolor{bggreen}PCB Merging & \cellcolor{top2bg}64.07 & \cellcolor{top2bg}23.00 & 29.00 & 64.33 & 81.31 & 21.40 & 32.20 & 31.20 & 42.00 & 20.80 & 33.96 & 55.22 & 3.65 & 83.54 & 57.45 \\
     \cellcolor{bggreen}DELLA & 24.27 & 0.50 & 9.00 & 34.94 & 45.83 & 18.70 & 28.99 & 26.00 & 37.41 & 31.20 & 42.96 & 23.11 & 6.53 & 69.79 & 37.08 \\
     \cellcolor{bggreen}SCE & 39.81 & 2.50 & 13.00 & \cellcolor{top1bg}66.79 & 54.45 & \cellcolor{top3bg}43.10 & \cellcolor{top1bg}62.86 & \cellcolor{top1bg}59.90 & \cellcolor{top1bg}71.26 & \cellcolor{top3bg}49.60 & \cellcolor{top2bg}62.86 & 47.04 & 9.75 & 83.54 & 58.66 \\
     \cellcolor{bggreen}WIDEN & 16.50 & 2.00 & 7.00 & 64.60 & 81.13 & 18.70 & 28.26 & 26.20 & 37.13 & 23.20 & 36.66 & 31.82 & 14.72 & 67.29 & 32.83 \\
     \cellcolor{bggreen}RAM+ & 58.25 & 12.50 & 7.00 & 61.92 & 79.49 & 22.20 & 34.30 & 28.00 & 41.51 & 25.60 & 42.53 & 46.29 & 1.67 & 79.37 & 50.46 \\
     \cellcolor{bggreen}TSV & 55.34 & 16.00 & 27.00 & 62.01 & 79.76 & 35.70 & 46.17 & 50.00 & 59.06 & 43.20 & 55.88 & 62.74 & 0.25 & 82.49 & 54.41 \\
     \cellcolor{bggreen}ISO-CTS & 0.97 & 0.00 & 0.00 & 0.00 & 0.18 & 0.00 & 0.00 & 0.00 & 0.00 & 0.00 & 0.00 & 4.77 & 16.68 & 0.00 & 0.00 \\
     \cellcolor{bggreen}IMPART & \cellcolor{top3bg}62.14 & 18.00 & 28.00 & 62.38 & 80.58 & 19.40 & 29.63 & 30.60 & 42.60 & 24.80 & 38.47 & 53.32 & 2.68 & 83.54 & \cellcolor{top1bg}62.31 \\
     \cellcolor{bggreen}WUDI & 60.19 & 19.00 & 26.00 & 60.44 & 78.31 & \cellcolor{top2bg}43.30 & \cellcolor{top3bg}54.69 & \cellcolor{top3bg}55.00 & \cellcolor{top3bg}65.87 & 48.00 & \cellcolor{top3bg}62.48 & \cellcolor{top3bg}66.23 & 0.85 & 80.63 & 55.62 \\
     \cellcolor{bggreen}DC-Merge & 57.84 & 16.00 & 26.00 & \cellcolor{top2bg}64.88 & 61.62 & 24.90 & 34.18 & 26.30 & 36.92 & 25.60 & 40.50 & 49.15 & 1.38 & 83.75 & 58.05 \\
     \cellcolor{bggreen}OrthoMerge-C & 58.25 & 16.50 & 23.00 & 62.57 & 79.04 & 29.10 & 40.34 & 40.80 & 52.38 & 36.00 & 50.65 & 58.43 & \cellcolor{top1bg}0.02 & 83.96 & 59.27 \\
     \cellcolor{bggreen}OrthoMerge-G & 40.78 & 18.00 & 25.00 & 62.38 & 81.03 & 27.40 & 38.16 & 37.20 & 47.50 & 37.60 & 50.84 & 56.07 & \cellcolor{top3bg}0.19 & \cellcolor{top1bg}86.66 & 57.45 \\
     \cellcolor{bgred}AdaMerging & 60.19 & 10.00 & 20.00 & 58.78 & 77.31 & 34.40 & 45.33 & 43.50 & 57.73 & \cellcolor{top3bg}49.60 & 62.29 & 57.38 & 1.11 & 81.66 & 41.03 \\
     \cellcolor{bgred}RegMean++ & 61.17 & 19.50 & 22.00 & \cellcolor{top3bg}64.66 & 78.95 & \cellcolor{top1bg}44.30 & \cellcolor{top2bg}56.38 & \cellcolor{top2bg}58.50 & \cellcolor{top2bg}68.23 & \cellcolor{top2bg}50.40 & \cellcolor{top1bg}65.34 & \cellcolor{top2bg}67.46 & 0.91 & 83.96 & 58.05 \\
     \cellcolor{bgred}CAT Merging & 61.17 & 14.00 & 24.00 & 63.59 & 79.58 & 25.60 & 37.36 & 36.00 & 48.32 & 38.40 & 53.53 & 57.51 & \cellcolor{top2bg}0.12 & \cellcolor{top2bg}85.21 & 58.05 \\
     \cellcolor{bgred}SLTA & 60.19 & 21.00 & \cellcolor{top3bg}31.00 & 62.38 & 80.31 & 16.60 & 27.21 & 23.00 & 34.81 & 20.80 & 33.98 & 50.36 & 5.96 & 82.91 & 57.75 \\
    \arrayrulecolor{black}\bottomrule
    \end{tabular}
    }
\end{table*}

\begin{table*}[t]
    \small
    \centering
    \caption{Performance comparison on the LoRA-trained Qwen3-30B-A3B-Think model. We select top-performing parameter-level methods to evaluate. For data- and parameter-level consolidation methods, the best, second-best, and third-best values are highlighted with \colorbox{top1bg}{darker}, \colorbox{top2bg}{medium}, and \colorbox{top3bg}{lighter} amber cells, respectively.}
    \label{tab:performance_comparison_30b_lora}
    \setlength{\tabcolsep}{3pt}
    \resizebox{\textwidth}{!}{%
    \begin{tabular}{l|ccc|cc|cccccc|cc|cc}
    \toprule
     % 第一行表头：Model占3行，Web/Doc/RAG跨列数调整
     \multirow{3}{*}{\textbf{Model}} & \multicolumn{3}{c|}{\textbf{Web}} & \multicolumn{2}{c|}{\textbf{Doc}} & \multicolumn{6}{c|}{\textbf{RAG}} & \multicolumn{2}{c|}{\textbf{Overall}} & \multicolumn{2}{c}{\textbf{OOD}}
     \\
     % 第二行表头：数据集名称
        & \textbf{GAIA} & \textbf{BC} & \textbf{BC-zh} & \textbf{MMBD} & \textbf{DocB} & \multicolumn{2}{c}{\textbf{HotPotQA}} & \multicolumn{2}{c}{\textbf{AmbigQA}} & \multicolumn{2}{c|}{\textbf{Bamboogle}} & \multirow{2}{*}{\textbf{G}$\uparrow$} & \multirow{2}{*}{\textbf{Imb.}$\downarrow$} & \textbf{AIME25} & \textbf{LCB} \\
     % 第三行表头：具体指标
        & \textit{Acc.} & \textit{Acc.} & \textit{Acc.} & \textit{Acc.} & \textit{Acc.} & \textit{EM} & \textit{F1} & \textit{EM} & \textit{F1} & \textit{EM} & \textit{F1} & & & \textit{Acc.} & \textit{Acc.} \\
    \midrule
    % 调整 multicolumn 宽度为 14
    \multicolumn{16}{l}{\textit{\textbf{Expert Agent}}} \\
    \arrayrulecolor{black!20}\midrule
     Web Agent ($\mathcal{D}_{\text{web}}$) & 57.28 & 13.00 & 16.00 & 61.46 & 78.22 & 32.20 & 40.99 & 32.60 & 38.15 & 32.80 & 40.84 & 54.80 & 0.46 & 72.91 & 43.47 \\
     Doc Agent ($\mathcal{D}_{\text{doc}}$) & 60.19 & 13.00 & 20.00 & 59.42 & 81.76 & 30.70 & 47.63 & 39.50 & 57.86 & 38.40 & 56.48 & 61.00 & 0.02 & 78.54 & 45.59 \\
     RAG Agent ($\mathcal{D}_{\text{rag}}$) & 35.92 & 9.00 & 17.00 & 48.96 & 72.95 & 44.30 & 56.27 & 58.50 & 69.18 & 56.00 & 69.31 & 50.10 & 4.47 & 81.87 & 51.67 \\
    \arrayrulecolor{black}\midrule
    \multicolumn{16}{l}{\textit{\textbf{Data-level Consolidation}}} \\
    \arrayrulecolor{black!20}\midrule
     Data Mixing ($\mathcal{D}_{\text{all}}$) & 57.28 & 16.00 & 21.00 & 62.11 & \cellcolor{top3bg}80.49 & \cellcolor{top2bg}44.30 & \cellcolor{top3bg}55.77 & \cellcolor{top3bg}57.30 & 66.16 & \cellcolor{top3bg}50.40 & \cellcolor{top3bg}63.68 & \cellcolor{top2bg}67.30 & 0.16 & 73.96 & 44.38 \\
    \arrayrulecolor{black}\midrule
    \multicolumn{16}{l}{\textit{\textbf{Parameter-level Consolidation}}} \\
    \arrayrulecolor{black!20}\midrule
     \cellcolor{bgblue}Average & \cellcolor{top1bg}61.17 & \cellcolor{top1bg}19.50 & \cellcolor{top1bg}25.00 & 60.81 & \cellcolor{top2bg}80.85 & 15.20 & 29.36 & 18.40 & 35.17 & 20.30 & 34.99 & 55.64 & 4.58 & 79.59 & 50.15 \\
     \cellcolor{bggreen}DARE & \cellcolor{top3bg}59.22 & 15.50 & \cellcolor{top3bg}23.00 & \cellcolor{top2bg}63.40 & \cellcolor{top3bg}80.49 & 13.60 & 28.01 & 16.70 & 34.30 & 8.80 & 27.48 & 51.96 & 5.58 & 78.56 & 48.94 \\
     \cellcolor{bggreen}TIES & 49.51 & 11.50 & 19.00 & 57.21 & 76.23 & 27.50 & 41.05 & 34.40 & 48.22 & 36.80 & 52.05 & 53.71 & \cellcolor{top2bg}0.07 & 75.63 & 45.59 \\
     \cellcolor{bggreen}SCE & 48.54 & 7.50 & 20.00 & 56.03 & 68.35 & \cellcolor{top3bg}43.60 & 55.33 & \cellcolor{top1bg}61.70 & \cellcolor{top1bg}72.53 & \cellcolor{top3bg}50.40 & \cellcolor{top2bg}63.85 & 55.23 & 2.27 & \cellcolor{top1bg}83.13 & \cellcolor{top3bg}58.66 \\
     \cellcolor{bggreen}RAM+ & 56.86 & 14.50 & 22.00 & 62.38 & \cellcolor{top2bg}80.85 & 22.80 & 39.40 & 19.70 & 33.53 & 25.60 & 42.40 & 56.14 & 1.83 & 78.12 & 49.54 \\
     \cellcolor{bggreen}TSV & 12.62 & 0.50 & 6.00 & 33.36 & 40.72 & 1.00 & 1.22 & 1.10 & 1.70 & 2.40 & 3.70 & 12.18 & 1.01 & 69.38 & 41.03 \\
     \cellcolor{bggreen}WUDI & 9.70 & 0.50 & 5.00 & 23.30 & 23.87 & 0.10 & 0.29 & 0.30 & 0.54 & 1.60 & 1.67 & 8.50 & 9.09 & 39.80 & 25.40 \\
     \cellcolor{bggreen}DC-Merge & 53.40 & 15.50 & 21.00 & 59.70 & 74.14 & 37.50 & 49.02 & 48.40 & 60.33 & 42.40 & 56.36 & 61.32 & 0.13 & \cellcolor{top3bg}81.88 & \cellcolor{top2bg}59.57 \\
     \cellcolor{bggreen}OrthoMerge-C & \cellcolor{top2bg}60.19 & 13.50 & 21.00 & 62.11 & 76.50 & 35.60 & 48.85 & 45.90 & 59.14 & 47.20 & 61.71 & 63.30 & \cellcolor{top1bg}0.03 & 81.66 & 55.93 \\
     \cellcolor{bggreen}OrthoMerge-G & 53.40 & \cellcolor{top2bg}18.50 & \cellcolor{top2bg}24.00 & \cellcolor{top3bg}63.21 & 79.04 & 15.80 & 29.13 & 25.60 & 40.24 & 16.80 & 32.50 & 55.08 & 3.63 & 77.91 & 49.54 \\
     \cellcolor{bgred}AdaMerging & 25.24 & 4.50 & 14.00 & 48.71 & 48.64 & 1.00 & 1.23 & 2.10 & 2.90 & 4.80 & 5.50 & 21.44 & 2.73 & 75.20 & 47.72 \\
     \cellcolor{bgred}RegMean++ & \cellcolor{top3bg}59.22 & \cellcolor{top3bg}17.50 & 21.00 & \cellcolor{top1bg}64.60 & \cellcolor{top1bg}82.03 & \cellcolor{top1bg}44.90 & \cellcolor{top2bg}56.02 & \cellcolor{top2bg}58.70 & \cellcolor{top2bg}68.53 & \cellcolor{top1bg}56.80 & \cellcolor{top1bg}66.45 & \cellcolor{top1bg}70.19 & \cellcolor{top3bg}0.12 & 79.39 & 49.85 \\
     \cellcolor{bgred}CAT Merging & 52.43 & 16.00 & 20.00 & 60.07 & 75.76 & 43.40 & \cellcolor{top1bg}56.22 & 55.00 & \cellcolor{top3bg}66.89 & \cellcolor{top2bg}52.00 & 63.62 & \cellcolor{top3bg}63.86 & 0.45 & \cellcolor{top2bg}82.72 & \cellcolor{top1bg}60.49 \\
     \cellcolor{bgred}SLTA & 58.25 & 17.00 & \cellcolor{top3bg}23.00 & 62.85 & 79.58 & 17.80 & 33.90 & 21.10 & 41.66 & 16.00 & 32.72 & 55.35 & 3.33 & 79.79 & 48.02 \\
    \arrayrulecolor{black}\bottomrule
    \end{tabular}
    }
\end{table*}

\begin{table}[t]
    \small
    \centering
    \caption{Out-of-domain capability retention across the three training scenarios. Each cell reports the \emph{absolute delta} accuracy vs.\ the backbone (Qwen3-30B-A3B-Think: AIME25$=81.87$, LCB$=57.14$;
    Qwen3-4B-Think: AIME25$=75.84$, LCB$=48.02$). A representative subset of methods is shown.}
    \label{tab:ood_retention}
    \setlength{\tabcolsep}{3pt}
    \resizebox{\columnwidth}{!}{%
    \begin{tabular}{l|cc|cc|cc}
    \toprule
    \multirow{2}{*}{\textbf{Method}}
      & \multicolumn{2}{c|}{\textbf{30B-A3B}}
      & \multicolumn{2}{c|}{\textbf{4B}}
      & \multicolumn{2}{c}{\textbf{30B-A3B (LoRA)}} \\
      & $\Delta$\textbf{AIME25} & $\Delta$\textbf{LCB}
      & $\Delta$\textbf{AIME25} & $\Delta$\textbf{LCB}
      & $\Delta$\textbf{AIME25} & $\Delta$\textbf{LCB} \\
    \midrule
        \multicolumn{7}{l}{\textit{\textbf{Data-level Consolidation}}} \\
    \arrayrulecolor{black!20}\midrule
        Data Mixing & \textcolor{red}{$-1.66$} & \textcolor{red}{$-1.82$} & \textcolor{red}{$-1.88$} & \textcolor{red}{$-11.85$} & \textcolor{red}{$-7.91$} & \textcolor{red}{$-12.76$} \\
    \arrayrulecolor{black}\midrule
    \multicolumn{7}{l}{\textit{\textbf{Parameter-level Consolidation (representative)}}} \\
    \arrayrulecolor{black!20}\midrule
        \cellcolor{bgblue}Average & \textcolor{green!60!black}{$+1.88$} & \textcolor{green!60!black}{$+4.26$} & \textcolor{green!60!black}{$+1.04$} & \textcolor{red}{$-3.95$} & \textcolor{red}{$-2.28$} & \textcolor{red}{$-6.99$} \\
        \cellcolor{bggreen}SCE & \textcolor{green!60!black}{$+1.67$} & \textcolor{green!60!black}{$+1.52$} & \textcolor{green!60!black}{$+3.74$} & \textcolor{red}{$-1.21$} & \textcolor{green!60!black}{$+1.26$} & \textcolor{green!60!black}{$+1.52$} \\
        \cellcolor{bggreen}DC-Merge & \textcolor{green!60!black}{$+1.88$} & \textcolor{green!60!black}{$+0.91$} & \textcolor{green!60!black}{$+4.78$} & \textcolor{green!60!black}{$+0.31$} & \textcolor{green!60!black}{$+0.01$} & \textcolor{green!60!black}{$+2.43$} \\
        \cellcolor{bggreen}OrthoMerge-C & \textcolor{green!60!black}{$+2.09$} & \textcolor{green!60!black}{$+2.13$} & \textcolor{green!60!black}{$+3.11$} & \textcolor{red}{$-0.30$} & \textcolor{red}{$-0.21$} & \textcolor{red}{$-1.21$} \\
        \cellcolor{bggreen}OrthoMerge-G & \textcolor{green!60!black}{$+4.79$} & \textcolor{green!60!black}{$+0.31$} & \textcolor{red}{$-0.84$} & \textcolor{green!60!black}{$+1.52$} & \textcolor{red}{$-3.96$} & \textcolor{red}{$-7.60$} \\
        \cellcolor{bggreen}WUDI & \textcolor{red}{$-1.24$} & \textcolor{red}{$-1.52$} & \textcolor{red}{$-1.47$} & \textcolor{red}{$-7.90$} & \textcolor{red}{$-42.07$} & \textcolor{red}{$-31.74$} \\
        \cellcolor{bgred}CAT Merging & \textcolor{green!60!black}{$+3.34$} & \textcolor{green!60!black}{$+0.91$} & \textcolor{green!60!black}{$+4.17$} & \textcolor{green!60!black}{$+0.92$} & \textcolor{green!60!black}{$+0.85$} & \textcolor{green!60!black}{$+3.35$} \\
        \cellcolor{bgred}RegMean++ & \textcolor{green!60!black}{$+2.09$} & \textcolor{green!60!black}{$+0.91$} & \textcolor{red}{$-1.68$} & \textcolor{red}{$-2.43$} & \textcolor{red}{$-2.48$} & \textcolor{red}{$-7.29$} \\
        \cellcolor{bgred}AdaMerging & \textcolor{red}{$-0.21$} & \textcolor{red}{$-16.11$} & \textcolor{red}{$-75.84$} & \textcolor{red}{$-36.77$} & \textcolor{red}{$-6.67$} & \textcolor{red}{$-9.42$} \\
    \arrayrulecolor{black}\bottomrule
    \end{tabular}
    }
\end{table}

\section{Empirical Study}
\label{sec:empirical}

We organize our empirical study around four research questions.
\textbf{(1) Performance}: can parameter-level consolidation match data-level mixing on aggregated in-domain performance?
\textbf{(2) Generalization}: how well do they retain out-of-domain capabilities such as math and code?
\textbf{(3) Failure Modes}: which training-scenario factors break which merging methods, and through what mechanism?
\textbf{(4) Mechanisms}: which design factors govern parameter-level merging's effectiveness, and how does consolidation reshape agent behavior?

\subsection{Performance: In-Domain Performance and Cross-Scenario Stability}
\label{sec:performance}

\noindent \textbf{Research Question 1.} \textit{How do parameter-level methods compare to data-level mixing in aggregated performance and cross-scenario stability?}

\begin{figure}[t]
  \centering
   \includegraphics[width=1.0\linewidth]{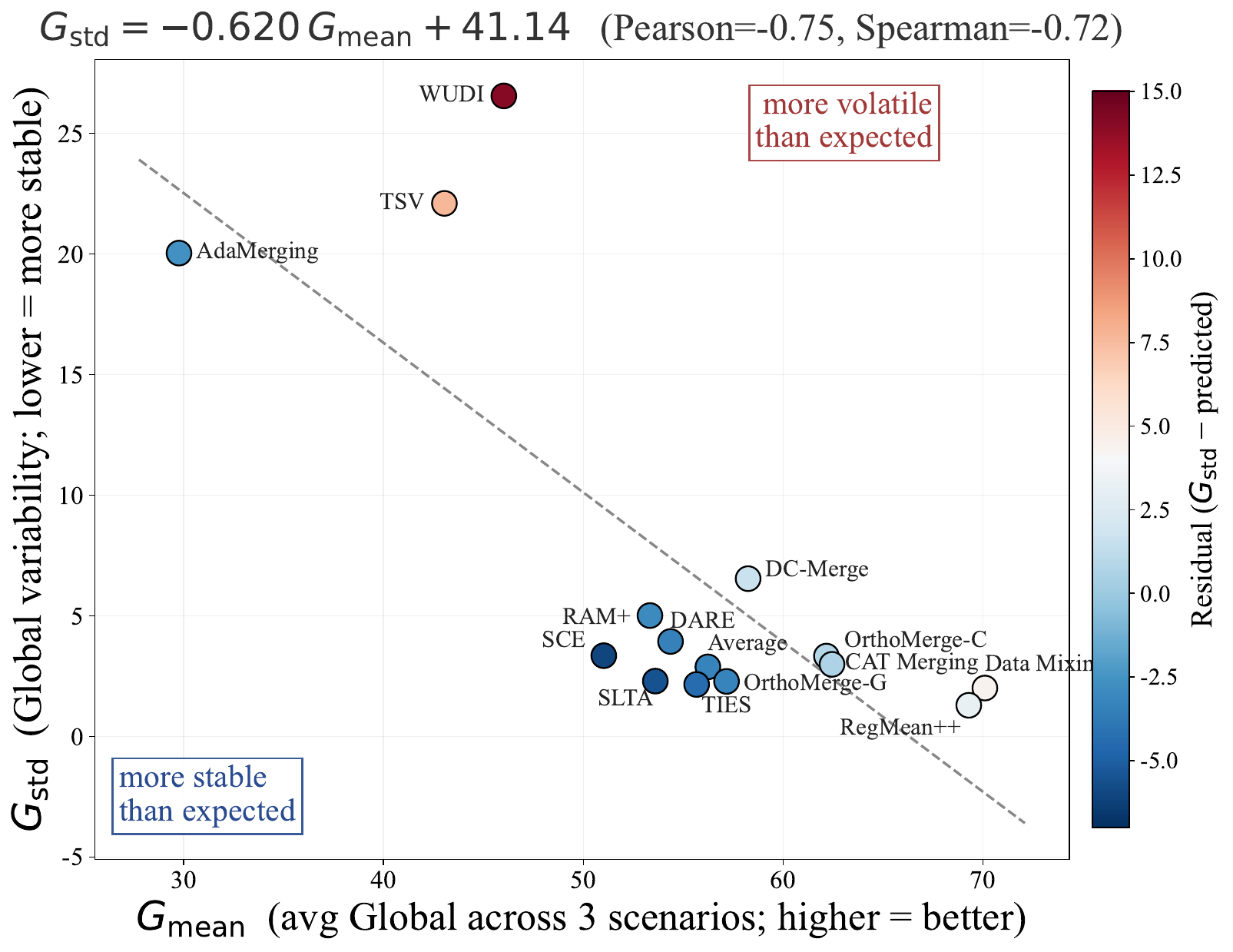}
   \caption{Cross-scenario quality-stability of representative methods. The $x$-axis is the mean Composite Score across the three training scenarios (Qwen3-30B-A3B-Think, Qwen3-4B-Think, and LoRA-trained Qwen3-30B-A3B-Think); the $y$-axis is its standard deviation (lower $=$ more stable). The dashed line is the OLS fit.}
   \label{fig:stability}
   \vspace{-3mm}
\end{figure}

\noindent \textbf{Results.}
\ding{182} \textbf{Aggregated co-champions.} Across the three training scenarios (Qwen3-30B-A3B-Think full-parameter, Qwen3-4B-Think full-parameter, and LoRA-trained Qwen3-30B-A3B-Think), data-level mixing remains a strong baseline. However, RegMean++ is closing the gap once we aggregate to the Composite Score $G$. RegMean++ and Data Mixing are the \emph{only} two methods ranking in the top-2 on all three scenarios, and they attain the lowest cross-scenario standard deviation. RegMean++ in particular uses a small amount of unlabelled data to obtain per-matrix output features and resolves both inter-model discrepancy and intra-model dependency in a closed form, reaching parity with data mixing. \textbf{Well-designed parameter-level consolidation can be comparable to data mixing.}
\ding{183} \textbf{Parameter-level merging is order-agnostic.} Parameter-level consolidation is invariant to the order in which experts are acquired. Data-level training, in contrast, depends critically on joint randomised mixing: applied sequentially on $\mathcal{D}_{\text{web}}$, $\mathcal{D}_{\text{doc}}$, and $\mathcal{D}_{\text{rag}}$, it suffers severe catastrophic forgetting---under a Web$\to$Doc$\to$RAG schedule, Web benchmark accuracy drops by $74.38\%$ relative to the post-Web stage and Doc benchmark accuracy by $59.27\%$ relative to the post-Doc stage (Appendix~\ref{app:practical_advantages}). This is a practical advantage of parameter-level merging when experts arrive incrementally.
\ding{184} \textbf{Stability is structurally tied to quality.} Across all three scenarios, $G_{\text{mean}}$ and $G_{\text{std}}$ are strongly negatively correlated (Pearson $r=-0.75$, Spearman $-0.72$, Figure~\ref{fig:stability}). Stability is therefore not an independent desideratum. It is a near-necessary condition for trustworthy merging, and new methods should be reported across $\geq 3$ scenarios to expose this axis. Appendix~\ref{app:four_class} additionally classifies all methods into four categories.

\subsection{Generalization: Out-of-Domain Capability Retention}
\label{sec:generalization}

\noindent \textbf{Research Question 2.} \textit{Do data-level and parameter-level methods preserve general capabilities outside the training distribution?}

\noindent \textbf{Results.}
\ding{182} \textbf{Data Mixing universally forgets OOD.} Data Mixing shows consistently negative $\Delta$AIME and $\Delta$LCB across all three scenarios. Moreover, the magnitude grows monotonically as the training setting becomes more constrained (full 30B-A3B $\to$ full 4B $\to$ LoRA 30B-A3B, combined $\Delta$ of $-3.48 \to -13.73 \to -20.67$). This reflects a direct trade of general capability for in-domain accuracy that is intrinsic to data-level multi-task fine-tuning, and it is amplified whenever the model has less spare capacity to absorb additional knowledge.
\ding{183} \textbf{A class of parameter-level methods structurally improves OOD.} CAT Merging, DC-Merge, and SCE achieve positive $\Delta$AIME and $\Delta$LCB on \emph{all} three scenarios. They operate at the task-vector level with explicit interference control via null-space projection, component decomposition, or sign consensus, suggesting a structural OOD-friendly inductive bias. RegMean++, although still net-negative on OOD, forgets substantially less than Data Mixing in every scenario, making it a more practical starting point for continual training or general-purpose deployment.

\subsection{Failure Modes Across Training Scenarios}
\label{sec:failure}

\noindent \textbf{Research Question 3.} \textit{Which training-scenario factors break which merging methods, and through what mechanism?}

\noindent \textbf{Results.}
\ding{182} \textbf{Subspace methods collapse under LoRA.} TSV and WUDI are strong on full-parameter trained Qwen3-30B-A3B-Think, but drop on LoRA scenarios. The mechanism is structural: LoRA updates are low-rank and reside in mutually orthogonal subspaces due to random initialization, lacking the spectral overlap that subspace alignment assumes. Instead of filtering noise, these methods truncate the distinct task-specific principal components essential for each domain, producing catastrophic information loss. ISO-CTS already fails for the same reason: its isotropic scaling suppresses the dominant principal encoding critical task knowledge while amplifying noise in the tail directions.
\ding{183} \textbf{Small-model fragility.} The 4B backbone exposes a different failure axis. \textit{(a)} \emph{AdaMerging collapses}: it uses training (rather than test, which is not available) samples as a proxy, causing unstable optimization and over-fitting. \textit{(b)} \emph{Single-domain experts erase out-of-domain capabilities at smaller models}: On Qwen3-4B-Think, Web and Doc Agents lose RAG ability. The effect largely disappears on Qwen3-30B-A3B-Think models, confirming that catastrophic forgetting in small models is exacerbated. Basic interpolation methods likewise exhibit larger 4B-vs-30B gaps on GAIA, consistent with the hypothesis that larger parameter spaces better absorb merging-induced interference~\cite{yadav2024matters}.

\subsection{Mechanisms: Design Factors and Behavioral Reshaping}
\label{sec:mechanisms}

\noindent \textbf{Research Question 4.} \textit{Which design factors govern parameter-level merging's effectiveness, and how does consolidation reshape agent behavior?}

\begin{figure}[t]
  \centering
   \includegraphics[width=\linewidth]{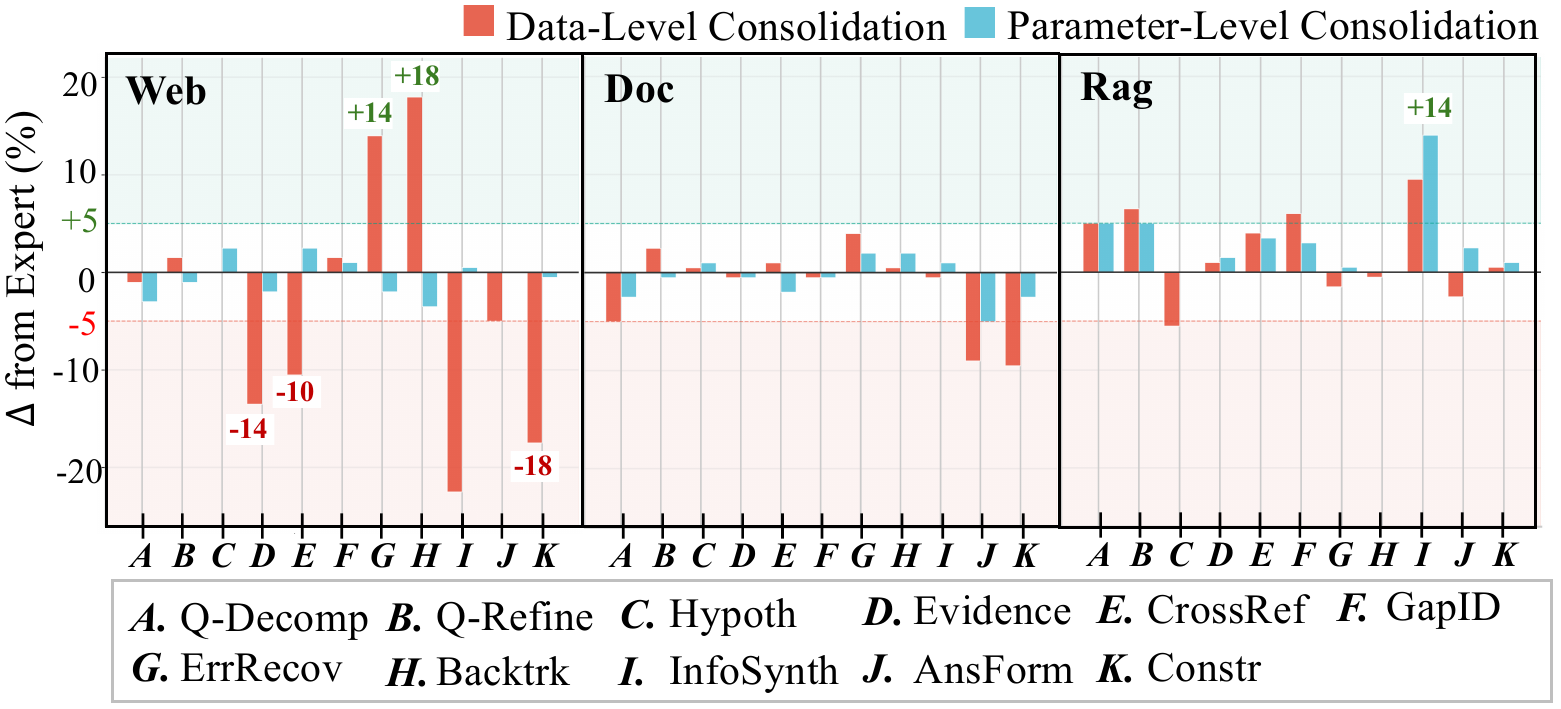}
   \caption{Differences of information-seeking behavior between consolidation methods and the expert agent across benchmarks on Qwen3-30B-A3B-Think. We select the top-performing parameter-level method, RegMean++, as representative. Results are reported across 11 information-seeking categories defined in Table~\ref{tab:behavior_categories}.}
   \label{fig:behavior}
\end{figure}

\noindent \textbf{Results.}
\ding{182} \textbf{Granularity.} Matrix-level methods consistently outperform model-level interpolation: expert agents exhibit layer-localised capabilities and heterogeneous parameter-update norms (Figure~\ref{fig:layer}), so coarse model-wise averaging dilutes them, whereas fine-grained matrix-wise optimisation preserves them~\cite{DBLP:journals/corr/abs-2508-06163}.
\ding{183} \textbf{Consensus-indicator informativeness.} The choice of consensus signal explains most of the spread among parameter-level methods. \textit{(a)} \emph{Basic interpolation} (Average, SLERP, MetaGPT) succeeds on homogeneous tasks with sphere/linear connectivity but drops sharply on heterogeneous domains (e.g., RAG). \textit{(b)} \emph{Heuristic sign consensus} (TIES, DELLA, Consensus TA, SCE) alleviates interference but is fragile. Consensus TA still lags substantially on RAG. DELLA and SCE substantially underperform on Web, and an inappropriate sign rule can trigger severe model collapse. \textit{(c)} \emph{Subspace alignment} (TSV, WUDI) is balanced on full-parameter settings but breaks on LoRA (\S\ref{sec:failure}). \textit{(d)} \emph{Data-dependent calibration} (RegMean++, CAT Merging) harmonises inter-model discrepancy and intra-model dependency, improving stability at the cost of forfeiting data independence. More informative consensus indicators yield more robust merges.
\ding{184} \textbf{Task affinity.} Web and Doc agents share tool-use patterns and information-seeking behaviors (Appendix~\ref{sec:app_tool_schema}) and can interchange under cross-domain evaluation, whereas RAG operates in a structurally distinct retrieval space. Merging agents with similar environments and behaviors is therefore more likely to recover expert-level performance~\cite{wang2025ui,team2025tongyi}.
\ding{185} \textbf{Behavior preservation.} Beyond accuracy, we annotate the information-seeking behavior of consolidated models against the expert agent on an 11-category schema (A--K) using Claude-Sonnet-4 (per-category deltas in Figure~\ref{fig:behavior}). Both data- and parameter-level consolidation alter agent behavior and diversity, but \emph{parameter-level method (RegMean++) preserves the expert agent's behavioral diversity substantially more than data-level mixing}. The tool-call analysis in Appendix~\ref{app:tool_calls} corroborates this via tool-call distributions.

\section{Future Outlook}
\label{sec:design_principles}

\noindent \textbf{Practical Recommendation.} Table~\ref{tab:method_selection} summarizes a condensation method selection guide for common deployment constraints.

\begin{table}[t]
    \small
    \centering
    \setlength{\tabcolsep}{3pt}
    \caption{Method selection guide by constraint.}
    \label{tab:method_selection}
    \resizebox{\columnwidth}{!}{%
    \begin{tabular}{l|p{0.30\columnwidth}|p{0.25\columnwidth}|p{0.27\columnwidth}}
    \toprule
    \textbf{Constraint} & \textbf{First choice} & \textbf{Backup} & \textbf{Avoid} \\
    \midrule
    Robust across all scenarios & RegMean++, Data Mixing & OrthoMerge-C, CAT Merging & WUDI/TSV (LoRA), AdaMerging (4B) \\
    \midrule
    LoRA-trained experts & RegMean++, Data Mixing, CAT Merging, OrthoMerge-C, DC-Merge & Doc Agent, RAM+ & WUDI, TSV, AdaMerging \\
    \midrule
    Small (4B) backbone & Data Mixing, RegMean++, OrthoMerge-C, CAT Merging, DC-Merge & WUDI, LiNeS, PCB Merging, OrthoMerge-G & Web/Doc Agent, CABS, AdaMerging \\
    \midrule
    30B-A3B full-parameter & RegMean++, Data Mixing, WUDI, TSV & Web/Doc Agent, OrthoMerge-C, CAT Merging & RAG Agent, WIDEN, SCE, DELLA, ISO-CTS, Consensus TA \\
    \midrule
    Preserve OOD math/code & CAT Merging, DC-Merge, SCE & OrthoMerge-C & Data Mixing \\
    \bottomrule
    \end{tabular}
    }
\end{table}

\noindent\textbf{Design Principles.} No existing method dominates across domains, scales, and training regimes. We distil four forward-looking principles for next-generation merging operators.

\textbf{1. Operate at matrix granularity.}
Agentic capabilities are layer-localised with heterogeneous update norms (Figure~\ref{fig:layer}). Future operators should default to per-matrix or per-module weighting rather than a single model-wise coefficient.

\textbf{2. Choose consensus indicators by informativeness, not convenience.}
Heuristic sign consensus is fragile, subspace alignment is regime-fragile, and full data calibration is robust but data-hungry. We advocate a \emph{cost-controlled middle ground}: lightweight statistics from a few-shot proxy set, with calibration cost (e.g., compute, privacy) treated as a design budget.

\textbf{3. Account for task affinity via adaptive coefficients.}
Web and Doc agents share tool-use schemas, whereas RAG is structurally distinct (Appendix~\ref{sec:app_tool_schema}), and uniform coefficients ignore this. Future methods should learn \emph{adaptive coefficients} that up-weight synergistic experts and balance orthogonal ones to maximise positive transfer.

\textbf{4. Respect the intrinsic geometry of parameter updates.}
Unlike fully fine-tuning, where task vectors often share overlapping singular value spectra, LoRA updates exhibit disjoint spectral signatures constrained by their low-rank initialization. Robust consolidation designs on LoRA must be adaptive to the updates that naturally reside in orthogonal subspaces.

% \textbf{5. Preserve behavioral diversity beyond accuracy.}
% Aggregated accuracy hides whether the merged agent still \emph{acts} like the expert. Future merging methods should explicitly evaluate---and ideally regularise---behavioral distributions (tool-call frequency, decomposition strategy, retrieval routines), with sparse-routing alternatives (e.g., MoE) when fidelity is at a premium.

\section{Conclusion}
We present the first systematic study of consolidating heterogeneous information-seeking agents across web, document, and knowledge-base environments, benchmarking 26 parameter-level methods against data mixing across three training scenarios. Well-designed parameter-level merging (e.g., RegMean++) matches data mixing at a fraction of its training cost and, unlike data mixing, structurally preserves out-of-domain capabilities. Effective merging requires finer granularity, informative consensus indicators, task-affinity-aware coefficients, and regime-adaptive geometry under LoRA. We hope this work provides foundation for future research on scalable and unified agentic models.

\section*{Limitations}
The training data of each expert are deliberately moderate in scale and held symmetric across the three domains to isolate consolidation effects from data-quantity confounds. Substantially skewed corpora could reveal regimes not visible in our experiments.

% Bibliography entries for the entire Anthology, followed by custom entries
%\bibliography{anthology,custom}
% Custom bibliography entries only
\bibliography{custom}

\appendix

\section{Discussion}

\noindent \textbf{Why agent consolidation is crucial?} 
Recent progress in agent reinforcement learning has largely focused on optimizing agents within individual task verticals, yet practical agentic systems must integrate behaviors learned under heterogeneous environments. 
While reinforcement learning improves vertical performance, it also induces strong environment-specific policy entanglement, making naive integration unstable and interference-prone. 
We view this work as an initial exploration of the agent consolidation problem, and hope that the observations and insights derived from our study can help inform and inspire subsequent work in this direction.
As a natural next step, we will extend this line of work to agentic reinforcement learning settings, where the interaction between policy optimization and consolidation, as well as the resulting increase in behavioral diversity, is expected to play a central role.

\section{Implementation Details}
\label{app:imple}
\begin{figure*}[t]
  \centering
   \includegraphics[width=0.95\linewidth]{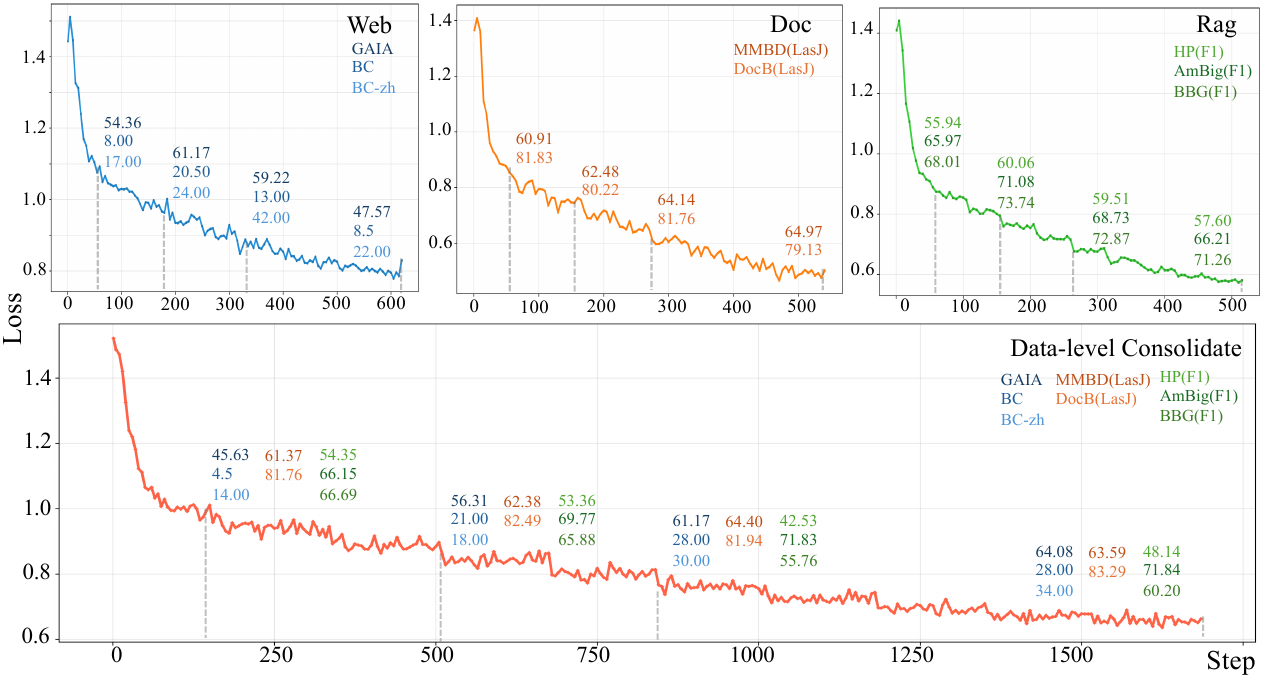}
   % \vspace{-3mm}
   \caption{Training loss curves and benchmark performance during training of Qwen3-30B-A3B-Think.}
   \label{fig:loss1}
   % \vspace{-5mm}
\end{figure*}

\begin{figure*}[t]
  \centering
   \includegraphics[width=0.95\linewidth]{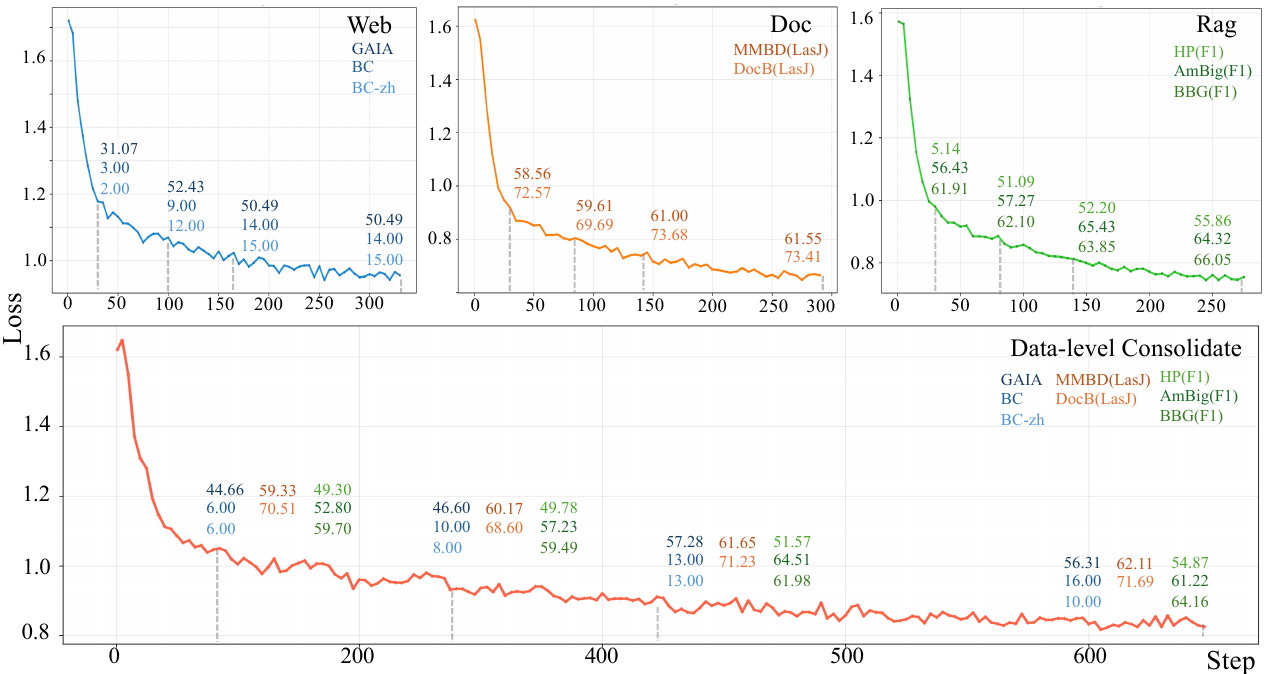}
   % \vspace{-3mm}
   \caption{Training loss curves and benchmark performance during training of Qwen3-4B-Think.}
   \label{fig:loss2}
   % \vspace{-5mm}
\end{figure*}

\begin{figure*}[t]
  \centering
   \includegraphics[width=0.95\linewidth]{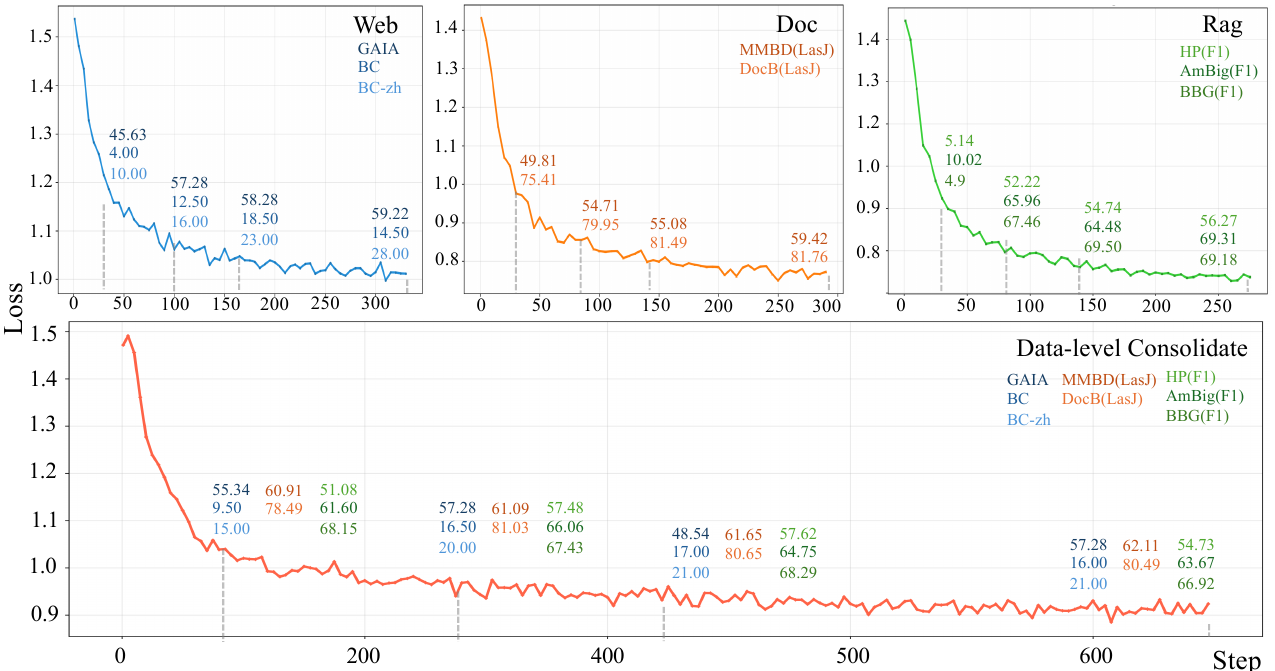}
   % \vspace{-3mm}
   \caption{Training loss curves and benchmark performance during training of Qwen3-30B-A3B-Think with LoRA.}
   \label{fig:loss3}
   % \vspace{-5mm}
\end{figure*}

We use Megatron-LM to finetune Qwen3-30B-A3B-Think\footnote{https://huggingface.co/Qwen/Qwen3-30B-A3B-Thinking-2507} and Qwen3-4B-Think\footnote{ https://huggingface.co/Qwen/Qwen3-4B-Thinking-2507}.
% Both models are trained with a context length of 128$k$. 
For LoRA training, we set LoRA rank to 8 and apply LoRA modules to all linear layers. 
The training loss curves and benchmark performance of trained Qwen3-30B-A3B-Think, Qwen3-4B-Think, and Qwen3-30B-A3B-Think with LoRA models are shown in Figures~\ref{fig:loss1}, \ref{fig:loss2}, and \ref{fig:loss3}, respectively.

We use the \textit{vLLM} framework~\cite{kwon2023efficient} for inference, with the temperature set to 0.6, the $top_p$ parameter set to 0.95, and the presence penalty set to 1.1.

\begin{figure*}[tbp]
  \centering
   \includegraphics[width=0.95\linewidth]{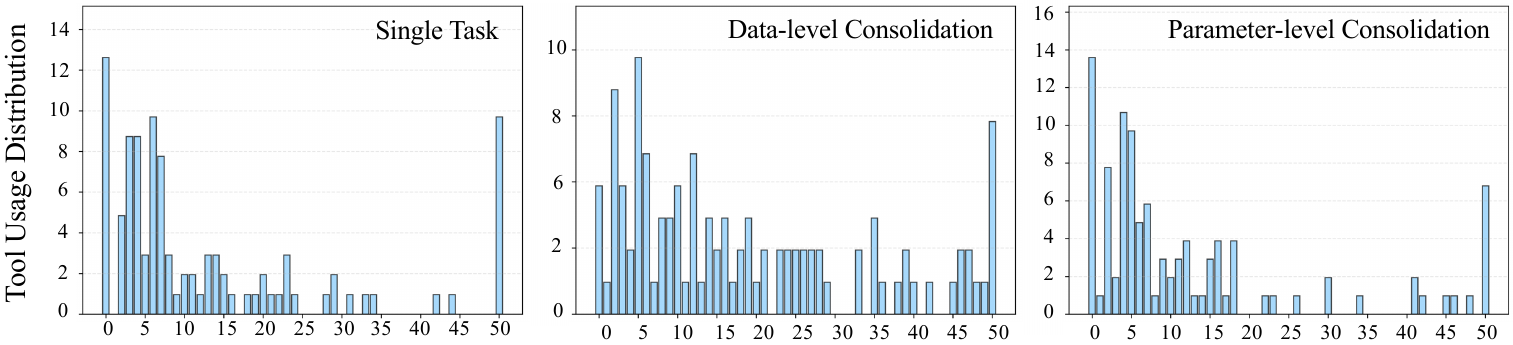}
   % \vspace{-3mm}
   \caption{Tool call distributions of different consolidation strategies across Web benchmarks.}
   \label{fig:tool_call_web}
   % \vspace{-5mm}
\end{figure*}

\begin{figure*}[tbp]
  \centering
   \includegraphics[width=0.95\linewidth]{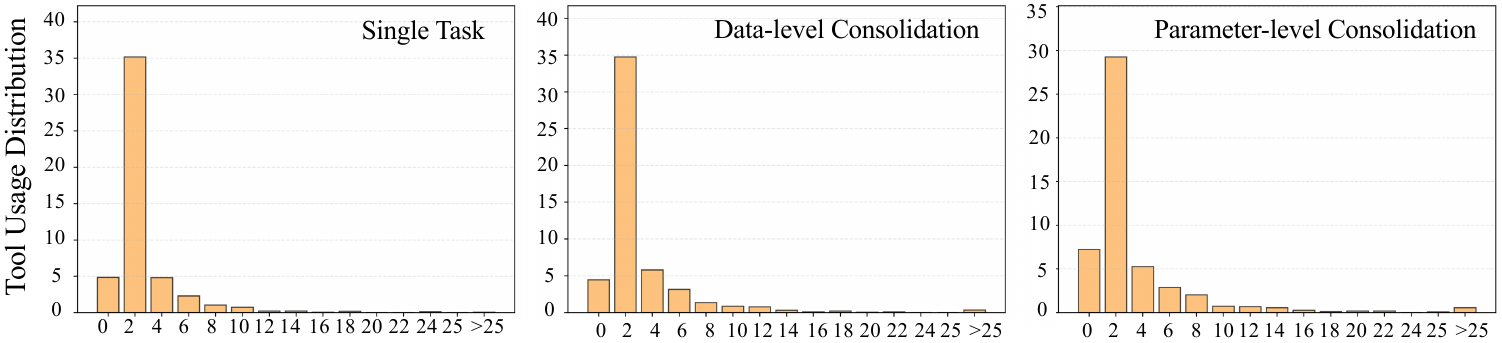}
   % \vspace{-3mm}
   \caption{Tool call distributions of different consolidation strategies across Doc benchmarks.}
   \label{fig:tool_call_doc}
   % \vspace{-5mm}
\end{figure*}

\begin{figure*}[tbp]
  \centering
   \includegraphics[width=0.95\linewidth]{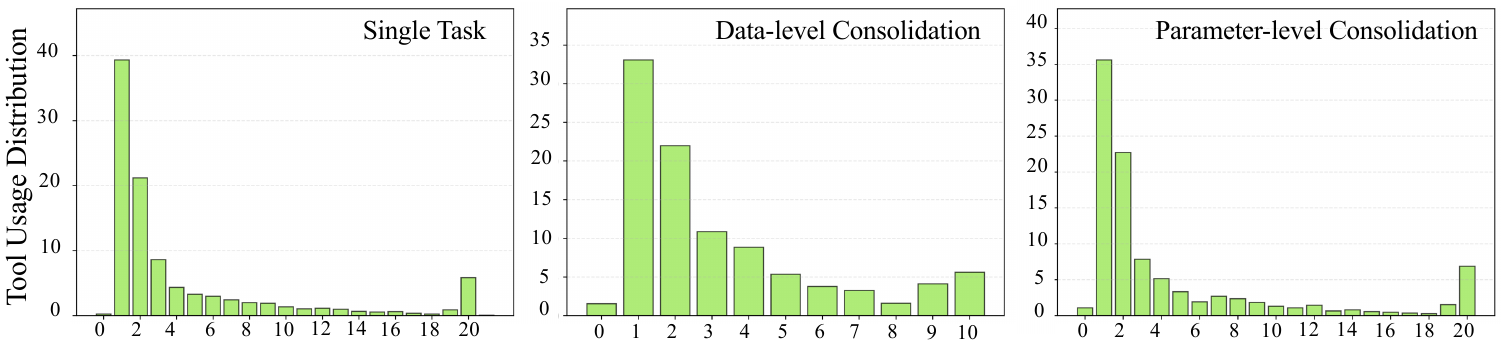}
   % \vspace{-3mm}
   \caption{Tool call distributions of different consolidation strategies across RAG benchmarks.}
   \label{fig:tool_call_rag}
   % \vspace{-5mm}
\end{figure*}

\section{Detailed Descriptions of Parameter-level Consolidation Methods}
\label{app:merge_methods}

Below, we provide brief descriptions of the 26 parameter-level consolidation methods evaluated in this work. The order is consistent with Table~\ref{tab:performance_comparison} and groups methods by the three categories in Table~\ref{tab:tax_merge_methods}.

\begin{description}[
    style=standard,
    leftmargin=0pt,
    itemindent=2em,
    nosep
]

% --- Group 1: Basic Interpolation ---
    \item[Average]~\cite{wortsman2022model} performs simple parameter averaging: $\theta^{(\text{merge})} = \sum_{i \in \mathcal{S}} \alpha^{(i)}\theta^{(i)}$. It assumes linear connectivity and aims to aggregate weights into a single centroid solution to improve generalization and robustness without increasing inference cost.

    \item[SLERP]~\cite{shoemake1985animating, ahmadian2024mix} merges models by interpolating parameters along a spherical path: $\theta^{(merge)} = \frac{\sin((1-t)\Omega)}{\sin(\Omega)} \theta_1 + \frac{\sin(t\Omega)}{\sin(\Omega)} \theta_2$. This accounts for the geometric structure of the high-dimensional parameter space.

    \item[MetaGPT]~\cite{zhou2024metagpt} solves for a regularized optimization problem for the scaling coefficients to merge.

    \item[LiNeS]~\cite{DBLP:conf/iclr/WangDFOFF25} applies depth-dependent scaling to parameter updates. It scales updates in deeper layers more aggressively while keeping shallow layers closer to pre-trained weights, reducing forgetting and interference.

% --- Group 2: Interference Resolution ---
    \item[DARE]~\cite{yu2024language} randomly drops $p\%$ parameter updates ($\boldsymbol{T} = \theta_{Expert} - \theta_{Base}$) on each expert model and rescales the remaining ones to reduce redundancy of task vectors and minimize interference between models.

    \item[Breadcrumbs]~\cite{davari2024model} constructs sparse masks by filtering out both insignificantly small perturbations and excessively large outliers from task vectors, retaining only the effective breadcrumbs.

    \item[TIES]~\cite{yadav2023ties} includes three steps: \textit{Trimming} insignificant redundant parameters, \textit{Electing} a unified sign direction based on majority voting (i.e., $s = \text{sign} \left( \sum_{i \in \mathcal{S}} \text{sign}(\theta^{(i)}) \right)$), and \textit{Merging} only the values that align with the elected direction.

    \item[Consensus TA]~\cite{wang2024localizing} constructs per-task masks to eliminate selfish weights. It filters these conflicting parameters to seek a consensus among task vectors.

    \item[TADrop]~\cite{DBLP:journals/corr/abs-2508-06163} is a distribution-aware sparsification method. It analyzes the weight distribution of each tensor to adaptively determine local sparsity ratios, preserving complex update patterns.

    \item[CABS]~\cite{DBLP:conf/icml/YangQ0LZ025} performs Conflict-Aware sparsification to prune overlapping and Balanced Sparsification with n:m pruning. This ensures a uniform distribution of retained weights across layers and prevents interference.

    \item[PCB Merging]~\cite{du2024parameter} scores parameter importance in task vectors based on intra- and inter-task balancing, and discards the bottom-ranked parameters.

    \item[DELLA]~\cite{deep2024della} drops parameter updates $\boldsymbol{T}$ based on magnitude and rescales the remaining ones. It prioritizes high-magnitude updates with task-critical information while randomly discarding low-magnitude updates.

    \item[SCE]~\cite{wan2025fusechat} includes three stages: \textit{Selecting} top $p\%$ elements with high variance; \textit{Calculating} matrix-level merging coefficients based on the sum of squares of these selected elements; \textit{Erasing} parameter updates with conflicting signs to eliminate interference.

    \item[WIDEN]~\cite{yu2024extend} disentangles each task vector into magnitude and direction components and re-weights them by per-parameter importance before fusion, alleviating magnitude conflicts that arise when task vectors carry highly heterogeneous scales.

    \item[RAM+]~\cite{yuan2026behavior} extends activation-aware re-basin merging with adaptive component selection: it first permutation-aligns expert models into a shared loss basin, then adaptively re-weights aligned components according to per-task activation statistics to mitigate residual interference.

    \item[TSV]~\cite{gargiulo2025task} leverages Singular Value Decomposition (SVD) to compress task vectors. It identifies the principal directions of weight updates and merges models in a low-rank subspace, retaining the most expressive components while reducing noise.

    \item[ISO-CTS]~\cite{DBLP:conf/icml/MarczakMCTB025} aligns task vectors by flattening their singular value spectrum. It explicitly models both a shared common subspace and task-specific subspaces to harmonize conflicts in the parameter space.

    \item[IMPART]~\cite{yang2025impart} utilizes SVD for importance-aware delta-sparsification. It dynamically adjusts the sparsity ratio for different singular vectors based on their contribution to the task.

    \item[WUDI]~\cite{DBLP:conf/icml/ChengXWZY25} directly minimizes a layer-wise interference objective. It identifies and mitigates specific components of task vectors that cause destructive interference, optimizing the merge without additional data.

    \item[DC-Merge]~\cite{zhang2026dc} decomposes each task vector into a shared common component and a task-specific residual component, and applies differential scaling to the two parts before aggregation, preserving consensus directions while attenuating task-specific noise.

    \item[OrthoMerge-C]~\cite{yang2026orthogonal} performs orthogonalization-based merging with a \emph{common}-subspace constraint: it projects each task vector onto a shared common subspace estimated from cross-task task vectors, suppressing components that lie outside the consensus direction.

    \item[OrthoMerge-G]~\cite{yang2026orthogonal} is the \emph{global} variant of OrthoMerge-C: it orthogonalizes each task vector against the union of all other task-vector directions, yielding stronger interference removal at the cost of more aggressive truncation of task-specific information.

% --- Group 3: Data-Driven Optimization ---
    \item[AdaMerging]~\cite{DBLP:conf/iclr/YangW00G0T24} automatically learns layer-wise merging coefficients by optimizing the entropy minimization objective $\min_{\boldsymbol{\lambda}} \mathbb{E}_{\boldsymbol{x} \sim \mathcal{D}_{\text{test}}} [\mathcal{H}(f(\boldsymbol{x}; \theta^{(base)} + \sum_{i} \boldsymbol{\lambda}_{i} \odot \boldsymbol{T}^{(i)}))]$ on unlabeled test data.

    \item[RegMean++]~\cite{DBLP:journals/corr/abs-2508-03121} enhances generalization by explicitly modeling intra- and cross-layer dependencies. It derives the closed-form solution $\boldsymbol{W}^{(merge)} = (\sum_{i} \hat{\boldsymbol{G}}_{i})^{-1} \sum_{i} \hat{\boldsymbol{G}}_{i} \boldsymbol{W}_{i}$, where $\hat{\boldsymbol{G}}_{i}$ represents the regularized inner-product matrix of input features propagated through the merged model.

    \item[CAT Merging]~\cite{DBLP:conf/icml/SunLGL25} resolves feature-level interference by projecting task vectors onto the null space of conflicting activations via the transformation $\boldsymbol{T}_{i} - \boldsymbol{T}_{i}\boldsymbol{B}_{k}\boldsymbol{B}_{k}^{\top}$. It selectively removes components aligned with the removal basis $\boldsymbol{B}_{k}$ that disrupt shared feature representations while preserving task-specific knowledge.

    \item[SLTA]~\cite{dai2025leveraging} performs \emph{Subspace Linear Task Arithmetic}: instead of operating in the full parameter space, it restricts task-vector arithmetic to a low-rank subspace identified from base-model activations, focusing the merge on representation-relevant directions and reducing off-manifold interference.
\end{description}

\section{Benchmarks}
\label{sec:app_benchmark}

This section provides a detailed overview of the benchmarks employed for Web, Document, and RAG agents.

\subsection{Web Agent Benchmarks}
\paragraph{GAIA~\cite{gaia}.} 
General AI Assistants (GAIA) is a benchmark designed to evaluate general-purpose AI systems on questions that are conceptually simple for humans but challenging for models due to the requirement for complex reasoning, tool usage, and multi-modality. Following the protocols in prior studies~\cite{webthinker,wu2025webdancer}, we focus on the text-only validation subset, comprising 103 instances. This subset isolates the agent's reasoning and browsing capabilities from visual processing factors. Performance is reported using accuracy.

\paragraph{BrowseComp~\cite{wei2025browsecomp}.} 
BrowseComp evaluates web agents on realistic user tasks that require interacting with real-world websites. The dataset emphasizes ``entangled information,'' where answers cannot be retrieved via simple queries and necessitate persistent navigation and multi-page information integration. Given the significant time and computational cost associated with live web browsing evaluation, we evaluate our method on a randomly sampled subset of 200 instances. We report success rates based on answer accuracy.

\paragraph{BrowseComp-zh~\cite{bc_zh}.} 
As the Chinese counterpart to BrowseComp, this dataset is constructed to reflect the unique characteristics of the Chinese internet ecosystem (e.g., distinct platform ecosystems and search behaviors). It tests the agent's robustness in non-English environments. Similar to BC, we utilize a randomly sampled subset of 100 instances for evaluation and report accuracy.

\subsection{Document Agent Benchmarks}
\paragraph{MMLongBenchDoc~\cite{ma2024mmlongbench}.} 
MMLongBenchDoc focuses on multimodal long-context document understanding. It features lengthy documents (averaging approximately 20k tokens, 47.5 pages) rich in layout elements such as charts, tables, and images across seven diverse domains.. A significant portion of the questions requires cross-page reasoning, testing the agent's ability to maintain long-term dependency and fuse multimodal information across extensive contexts.

\paragraph{DocBench~\cite{zou2025docbench}.} 
DocBench consists of 229 real-world documents and 1,082 questions, designed to assess the robustness of document reading systems across five domains and four major question types, providing a holistic view of an agent's document processing capabilities.

\subsection{RAG Agent Benchmarks}
\paragraph{HotPotQA~\cite{yang2018hotpotqa}.} 
HotPotQA is specifically designed to test multi-hop reasoning. Answering questions requires the agent to find and reason over multiple supporting documents to derive the correct answer, challenging the agent's ability to decompose complex queries.

\paragraph{AmbigQA~\cite{min2020ambigqa}.} 
AmbigQA addresses the challenge of ambiguity in open-domain questions. The agent must handle queries with multiple potential answers by retrieving diverse evidence or disambiguating the context, thereby testing the precision and coverage of the retrieval process.

\paragraph{Bamboogle~\cite{press2023measuring}.} 
Bamboogle consists of questions where the answer cannot be found on the first page of standard search engine results. This dataset evaluates the agent's resilience and its ability to perform multi-step retrieval when direct search fails.

\subsection{Out-of-Domain Benchmarks}
\label{app:ood_benchmarks}
To assess whether the agent-consolidated model still retains the broader reasoning capabilities of its backbone, we evaluate it on two benchmarks that lie completely outside the Web, Document, and RAG training distributions. Both probe long-horizon symbolic reasoning, one in natural language with formal math, and one in executable code, which our training data never directly targets.

\paragraph{AIME 2025~\cite{aime2025}.}
The American Invitational Mathematics Examination is a long-running selective high-school olympiad administered by the Mathematical Association of America. The 2025 edition consists of two contests (AIME~I and AIME~II), each comprising 15 problems whose answers are integers in $[0, 999]$, totalling 30 problems that cover algebra, combinatorics, number theory, and geometry. The integer-answer format eliminates judging ambiguity, allowing strict exact-match scoring; problems require multi-step symbolic manipulation rather than rote computation, making AIME a clean probe of out-of-distribution mathematical reasoning. We evaluate under chain-of-thought decoding and report accuracy (percentage of correctly answered problems averaged over both contests).

\paragraph{LiveCodeBench~\cite{jain2024livecodebench}.}
LiveCodeBench (LCB) is a holistic, contamination-free coding benchmark that continuously aggregates new problems from competitive-programming platforms (LeetCode, AtCoder, and Codeforces) and re-releases the test suite as periodically updated time slices. Each problem ships with a hidden test bank that judges Pass@1 by executing the generated code against unseen inputs. We adopt the \texttt{2408-2502} time slice, which restricts evaluation to problems released between August 2024 and February 2025; this time-windowed protocol is designed to minimise overlap with model pre-training corpora and provides a tighter measurement of genuine code-generation skill than static benchmarks. We report Pass@1 accuracy under greedy decoding.

\section{Agents}
\label{app:agents}
\subsection{Web Agent}
The web agent employs two types of tools, following previous work~\cite{wu2025webdancer,tao2025webshaper}: 
\textit{Search} and \textit{Visit}:
\begin{itemize}[leftmargin=1.5em, itemsep=2pt, topsep=1pt, parsep=0pt]
    \item \textbf{\textit{Search}} is used to retrieve information via the Google search engine. 
    Its inputs are search queries, and it supports issuing multiple queries in parallel. 
    For each query, the tool returns the top-10 results, where each result includes a title, a snippet, and the corresponding URL.
    \item \textbf{\textit{Visit}} is used to access and process specific web pages. 
    The input consists of a set of urls, each paired with a dedicated visit goal.
    The full content of each page is first retrieved using Jina, after which a summarization model extracts goal-relevant information. In this work, we use gpt-oss-120b~\cite{agarwal2025gpt} as the summarization model.
\end{itemize}

\subsection{Doc Agent}

The document agent employs two types of tools, following previous work~\cite{zhang2026docdancer}:
\begin{itemize}[leftmargin=1.5em, itemsep=2pt, topsep=1pt, parsep=0pt]
    \item \textbf{\textit{Search.}} is used to conduct keyword-based full-text search over the given documents. Its inputs are search keywords, and it returns the corresponding section IDs, page numbers, and surrounding text snippets for each match. A visible window is used to constrain the snippet length for efficient localization. This tool provides the agent with global textual signals for guiding subsequent information access.
    \item \textbf{\textit{Read.}} is used to access and process specific document sections. The input consists of a goal and a set of section IDs. For each section, the tool first retrieves local textual information, consisting of all text within the section, and local visual information, consisting of images and tables within the section, together with a page-level screenshot that captures the full layout of the page containing the section. Subsequently, a multimodal summarization model $M_m$ is used as an auxiliary reader to jointly integrate textual and visual inputs and return consolidated goal-relevant content.
\end{itemize}

\subsection{RAG Agent}
The RAG agent employs a dense retrieval tool:
\begin{itemize}[leftmargin=1.5em, itemsep=2pt, topsep=1pt, parsep=0pt]
    \item \textbf{\textit{Dense Retrieval.}} is used to retrieve relevant documents from a knowledge base, Wikipedia. The parameters include a query and a top-$k$ value, representing the search string and the maximum number of relevant documents to return, respectively. 
    The tool encodes the query using a pretrained text embedding model\footnote{https://github.com/facebookresearch/DPR}~\cite{karpukhin2020dense} and computes similarity scores between the query and documents indexed in the KB. It returns documents whose similarity scores exceed a threshold $\tau$, while ensuring that the number of returned documents does not exceed $k$.
\end{itemize}

\subsection{Tool Schema}
\label{sec:app_tool_schema}

This section details the tool schemas provided to the agent. 
The specific JSON structures defining the tools for the Web agent, Doc agent, and RAG agent are shown in Figures~\ref{fig:web_tool_schema}, ~\ref{fig:doc_tool_schema}, and~\ref{fig:rag_tool_schema}.

\lstset{
  literate={"}{{\texttt{"}}}1
}

\begin{figure*}[t]
    \definecolor{solarizedBase}{RGB}{242, 245, 250}
    \definecolor{solarizedFrame}{RGB}{081, 132, 178}
    \definecolor{codeGreen}{RGB}{133, 153, 0}
    \definecolor{codeCyan}{RGB}{42, 161, 152}

    \lstset{
        basicstyle=\ttfamily\footnotesize,
        breaklines=true,
        columns=fullflexible,
        stringstyle=\color{codeCyan},
        keywordstyle=\color{codeGreen},
        upquote=true,
    }

    \begin{tcolorbox}[
        title={Tool Schemas for Web Agent},
        colback=solarizedBase!20,
        colframe=solarizedFrame,
        coltitle=white,
        fonttitle=\bfseries,
        width=\textwidth
    ]

    \textbf{\textit{Search}}
\begin{lstlisting}
{
  "type": "function",
  "function": {
    "name": "search",
    "description": "Performs batched web searches: supply an array 'query'; the tool retrieves the top 10 results for each query in one call.",
    "parameters": {
      "type": "object",
      "properties": {
        "query": {
          "type": "array",
          "items": {
            "type": "string"
          },
          "description": "Array of query strings. Include multiple complementary search queries in a single call."
        }
      },
      "required": ["query"]
    }
  }
}
\end{lstlisting}

    \textbf{\textit{Visit}}
\begin{lstlisting}
{
  "type": "function",
  "function": {
    "name": "visit",
    "description": "Visit webpage(s) and return the summary of the content.",
    "parameters": {
      "type": "object",
      "properties": {
        "url": {
          "type": ["string", "array"],
          "items": {
            "type": "string"
          },
          "minItems": 1,
          "description": "The URL(s) of the webpage(s) to visit. Can be a single URL or an array of URLs."
        },
        "goal": {
          "type": "string",
          "description": "The goal of the visit for webpage(s)."
        }
      },
      "required": ["url", "goal"]
    }
  }
}
\end{lstlisting}

    \end{tcolorbox}
    \caption{Tool schema for web agent: \textit{Search} and \textit{Visit}.}
    \label{fig:web_tool_schema}
\end{figure*}

\begin{figure*}[t] 
    \definecolor{solarizedBase}{RGB}{242, 245, 250} % 米黄色背景
    \definecolor{solarizedFrame}{RGB}{081, 132, 178}  % 黄铜色/深黄色
    \definecolor{codeGreen}{RGB}{133, 153, 0}
    \definecolor{codeCyan}{RGB}{42, 161, 152}

    \lstset{
        basicstyle=\ttfamily\footnotesize,
        breaklines=true,
        columns=fullflexible,
        stringstyle=\color{codeCyan},
        keywordstyle=\color{codeGreen},
        upquote=true,
    }

    \begin{tcolorbox}[
        title={Tool Schemas for Doc Agent},
        colback=solarizedBase!20,      % 暖色背景
        colframe=solarizedFrame,    % 暖色边框
        coltitle=white,
        fonttitle=\bfseries,
        width=\textwidth 
    ]
    \textbf{\textit{Search}}
    \begin{lstlisting}
{
    "type": "function",
    "function": {
        "name": "search",
        "description": "Find and extract all paragraphs and sections where any of the provided search terms appear",
        "parameters": {
            "type": "object",
            "properties": {
                "keywords": {
                    "type": "array",
                    "items": {
                        "type": "string"
                    },
                    "description": "A list of query keywords for searching"
                }
            },
            "required": ["keywords"]
        }
    }
}
\end{lstlisting}
\textbf{\textit{Read}}
\begin{lstlisting}
{
    "type": "function",
    "function": {
        "name": "read",
        "description": "Read multiple sections by section IDs and extract useful information from all content contained in those sections, including both visual elements and textual elements.",
        "parameters": {
            "type": "object",
            "properties": {
                "section_ids": {
                    "type": "array",
                    "items": {
                        "type": "string"
                    },
                    "description": "A list of section IDs to read from the document"
                },
                "goal": {
                    "type": "string",
                    "description": "The user goal that guides what useful information should be extracted from the selected sections"
                }
            },
            "required": ["section_ids", "goal"]
        }
    }
}
    \end{lstlisting}
    \end{tcolorbox}
    \caption{Tool schema for doc agent: \textit{Search} and \textit{Read}.}
    \label{fig:doc_tool_schema}
\end{figure*} 

\begin{figure*}[t]
    \definecolor{solarizedBase}{RGB}{242, 245, 250}
    \definecolor{solarizedFrame}{RGB}{081, 132, 178}
    \definecolor{codeGreen}{RGB}{133, 153, 0}
    \definecolor{codeCyan}{RGB}{42, 161, 152}

    \lstset{
        basicstyle=\ttfamily\footnotesize,
        breaklines=true,
        columns=fullflexible,
        stringstyle=\color{codeCyan},
        keywordstyle=\color{codeGreen},
        upquote=true,
    }

    \begin{tcolorbox}[
        title={Tool Schema for RAG Agent},
        colback=solarizedBase!20,
        colframe=solarizedFrame,
        coltitle=white,
        fonttitle=\bfseries,
        width=\textwidth
    ]

    \textbf{\textit{Dense Retrieval}}
\begin{lstlisting}
{
  "type": "function",
  "function": {
    "name": "dense_retrival",
    "description": "Semantic vector search over the knowledge base. Falls back to configured top_k or 5.",
    "parameters": {
      "type": "object",
      "properties": {
        "query": {
          "type": "string",
          "minLength": 1,
          "description": "Natural language question or statement to retrieve against the knowledge base."
        },
        "top_k": {
          "type": "integer",
          "minimum": 1,
          "description": "Optional override for number of retrieved results."
        }
      },
      "required": ["query"]
    }
  }
}
\end{lstlisting}

    \end{tcolorbox}
    \caption{Tool schema for RAG agent: dense semantic retrieval over a vectorized knowledge base.}
    \label{fig:rag_tool_schema}
\end{figure*}

\section{Results on Qwen3-4B-Think}
We report the expert agent, data-level consolidation, and parameter-level consolidation performance
comparison results on Qwen3-4B-Think model in Table~\ref{tab:performance_comparison_4b}.

\section{Practical Advantages of Parameter-level Consolidation}
\label{app:practical_advantages}

Although both data-level mixing and parameter-level consolidation can produce a single multi-capability agent, parameter-level merging offers three deployment-oriented advantages that are difficult to recover with data-level mixing. We summarise them here as a complement to the in-domain and OOD comparisons in \S\ref{sec:performance}-\S\ref{sec:generalization}.

\paragraph{1. Modular continual learning.} Data-level consolidation requires re-training on the union of all task corpora whenever a new domain arrives---approximately $1{,}960$ A100 GPU\,h for our three-task setting (Web, Doc, RAG), and the cost scales linearly with the number of incorporated tasks. Parameter-level consolidation instead maintains an independent expert per task and combines arbitrary subsets on demand. In our pipeline, most merges take $<\!20$ GPU\,h, a $\sim\!10^2\times$ speed-up. When experts must be acquired sequentially (rather than mixed jointly), data-level fine-tuning further suffers from catastrophic forgetting. We quantify this effect in Table~\ref{tab:sequential_forgetting} below. We sequentially fine-tune the Qwen3-30B-A3B-Think backbone in the order Web $\to$ Doc $\to$ RAG. Because Web and Doc share tool-use patterns and information-seeking behaviors (Appendix~\ref{sec:app_tool_schema}), training on Doc after Web yields a mild positive transfer on the Web benchmarks. The subsequent RAG stage, however, severely erases the previously acquired Web and Doc capabilities: averaged across the three Web benchmarks, accuracy collapses by $74.38\%$ relative to the post-Web stage; averaged across the two Doc benchmarks, by $59.27\%$ relative to the post-Doc stage. The final model essentially regresses to a single-domain RAG expert. In contrast, parameter-level merging is order-agnostic by construction (\S\ref{sec:performance}, \ding{183}): the consolidated weights are a symmetric function of $\{\theta_{\text{web}}, \theta_{\text{doc}}, \theta_{\text{rag}}\}$ and exhibit no schedule sensitivity, which justifies its preference whenever experts arrive incrementally or training cannot be repeated.

\paragraph{2. Data sovereignty.} Parameter-level merging is the only viable option when source datasets cannot be pooled because of privacy regulations (e.g., medical or enterprise records), licensing restrictions, or jurisdictional constraints. It is also the natural integration path when only \emph{pre-trained expert checkpoints} are available---e.g., when consolidating community-released agents whose original training data are no longer accessible.

\paragraph{3. Better OOD knowledge retention.} As established in \S\ref{sec:generalization} and Table~\ref{tab:ood_retention}, several parameter-level methods achieve positive $\Delta$AIME and $\Delta$LCB on \emph{all} three training scenarios, whereas Data Mixing forgets math and code capabilities universally, with the magnitude growing as the training budget tightens (full 30B-A3B $\to$ full 4B $\to$ LoRA 30B-A3B, combined $\Delta$ of $-3.48 \to -13.73 \to -20.67$).

Together, these three properties enable concrete deployment scenarios that are otherwise impractical: \textit{(i)} privacy-preserving or federated consolidation across distributed expert holders; \textit{(ii)} integration of pre-existing community agents without re-collecting their training data; and \textit{(iii)} rapid prototyping by merging a curated subset of experts on demand.

\begin{table*}[h]
    \small
    \centering
    \caption{Sequential fine-tuning of Qwen3-30B-A3B-Think under the Web $\to$ Doc $\to$ RAG schedule. Each row reports per-benchmark accuracy at the end of the corresponding training stage (RAG benchmarks are EM\,/\,F1). Continual training on the next domain steadily erases previously acquired capabilities, with the final RAG stage causing a $74.38\%$ Web drop and a $59.27\%$ Doc drop relative to the prior stage-empirical evidence of catastrophic forgetting in data-level continual fine-tuning.}
    \label{tab:sequential_forgetting}
    \setlength{\tabcolsep}{3pt}
    \resizebox{0.8\textwidth}{!}{%
    \begin{tabular}{l|ccc|cc|ccc}
    \toprule
    \multirow{2}{*}{\textbf{Stage}}
      & \multicolumn{3}{c|}{\textbf{Web}}
      & \multicolumn{2}{c|}{\textbf{Doc}}
      & \multicolumn{3}{c}{\textbf{RAG} (EM\,/\,F1)} \\
      & \textbf{GAIA} & \textbf{BC} & \textbf{BC-zh}
      & \textbf{MMBD} & \textbf{DocB}
      & \textbf{HotPotQA} & \textbf{AmbigQA} & \textbf{Bamboogle} \\
    \midrule
    Web                                & 62.14 & 21.00 & 24.00 & 55.82 & 79.58 & 34.10\,/\,50.09 & 44.40\,/\,60.87 & 44.00\,/\,61.60 \\
    Web $\to$ Doc                      & 65.64 & 22.00 & 30.00 & 64.42 & 80.76 & 35.20\,/\,51.45 & 48.00\,/\,64.83 & 40.10\,/\,58.53 \\
    Web $\to$ Doc $\to$ RAG            & 19.41 & \phantom{0}1.00 & \phantom{0}7.00 & 27.91 & 31.22 & 43.60\,/\,54.77 & 54.40\,/\,65.42 & 59.10\,/\,68.50 \\
    \bottomrule
    \end{tabular}
    }
\end{table*}

\begin{figure}[t]
  \centering
   \includegraphics[width=1.0\linewidth]{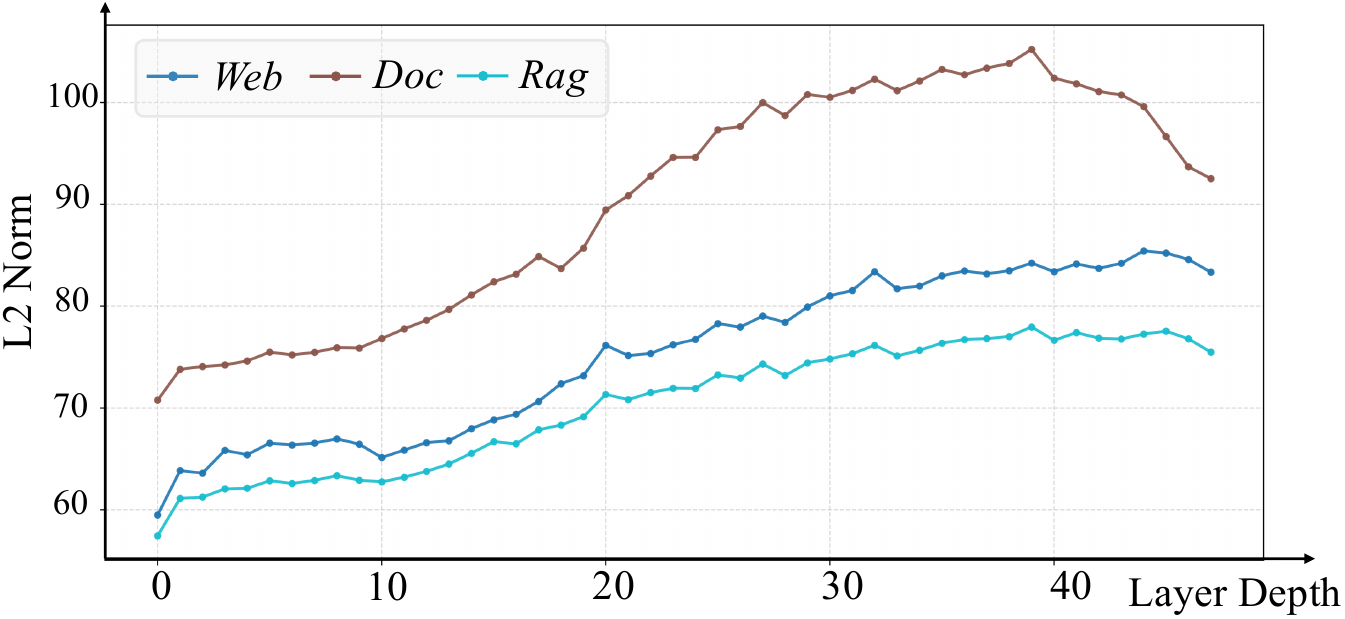}
   \caption{Layer-wise L2 norm of parameter updates of expert agents for Web, Doc, and RAG agents across model depth.}
   \label{fig:layer}
   \vspace{-3mm}
\end{figure}

\begin{figure*}[t]
   \centering
   \includegraphics[width=1.0\linewidth]{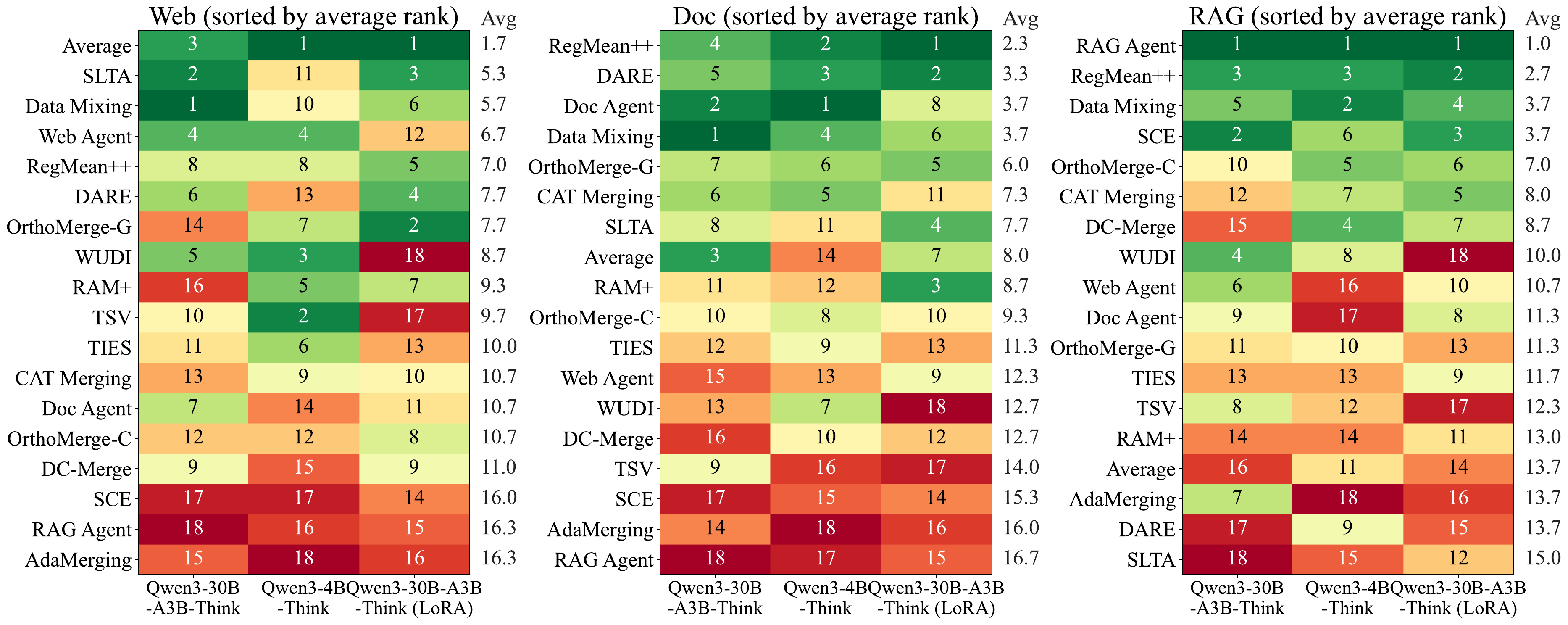}
   \caption{Per-capability rankings of merging methods across the three training scenarios (Qwen3-30B-A3B-Think, Qwen3-4B-Think, and LoRA-trained Qwen3-30B-A3B-Think). Each cell encodes the method's rank for that capability in that scenario (greener $=$ better rank). Left: Web; centre: Doc; right: RAG. Methods within each panel are sorted by their mean rank across scenarios.}
   \label{fig:single_capability}
\end{figure*}

\section{Comparison on Tool Call}
\label{app:tool_calls}
Figures~\ref{fig:tool_call_web},~\ref{fig:tool_call_doc}, and~\ref{fig:tool_call_rag} respectively illustrate the tool call distributions of the expert agent, data-level consolidation, and top-performing parameter-level consolidation on Web benchmarks, Doc benchmarks, and RAG benchmarks.
Parameter-level consolidation exhibits a tool-call distribution that is closer to data-level consolidation and the expert agent.

\begin{table*}[tbp]
    \small
    \centering
    \caption{Performance comparison with Qwen3-4B-Think model as backbone. \textbf{G} and \textbf{Imb.} are the Composite Score and Imbalance Score describe the overall performance and task imbalance on three tasks. For data- and parameter-level consolidation, the best, second-best, and third-best values are highlighted with \colorbox{top1bg}{darker}, \colorbox{top2bg}{medium}, and \colorbox{top3bg}{lighter} amber cells, respectively.}
    \label{tab:performance_comparison_4b}
    \setlength{\tabcolsep}{3pt}
    \resizebox{\textwidth}{!}{%
    \begin{tabular}{l|ccc|cc|cccccc|cc|cc}
    \toprule
     \multirow{3}{*}{\textbf{Model}} & \multicolumn{3}{c|}{\textbf{Web}} & \multicolumn{2}{c|}{\textbf{Doc}} & \multicolumn{6}{c|}{\textbf{RAG}} & \multicolumn{2}{c|}{\textbf{Overall}} & \multicolumn{2}{c}{\textbf{OOD}}
     \\
     % 第二行表头：数据集名称
        & \textbf{GAIA} & \textbf{BC} & \textbf{BC-zh} & \textbf{MMBD} & \textbf{DocB} & \multicolumn{2}{c}{\textbf{HotPotQA}} & \multicolumn{2}{c}{\textbf{AmbigQA}} & \multicolumn{2}{c|}{\textbf{Bamboogle}} & \multirow{2}{*}{\textbf{G}$\uparrow$} & \multirow{2}{*}{\textbf{Imb.}$\downarrow$} & \textbf{AIME25} & \textbf{LCB} \\
     % 第三行表头：具体指标
        & \textit{Acc.} & \textit{Acc.} & \textit{Acc.} & \textit{Acc.} & \textit{Acc.} & \textit{EM} & \textit{F1} & \textit{EM} & \textit{F1} & \textit{EM} & \textit{F1} & & & \textit{Acc.} & \textit{Acc.} \\
    \midrule
    \multicolumn{16}{l}{\textit{\textbf{Expert Agent}}} \\
    \arrayrulecolor{black!20}\midrule
     Web Agent ($\mathcal{D}_{\text{web}}$) & 50.49 & 14.00 & 15.00 & 57.39 & 69.98 & 1.60 & 2.80 & 4.60 & 7.43 & 3.20 & 5.15 & 43.55 & 11.68 & 72.08 & 39.82 \\
     Doc Agent ($\mathcal{D}_{\text{doc}}$) & 46.60 & 7.50 & 15.00 & 61.00 & 73.68 & 1.40 & 3.10 & 2.80 & 4.00 & 2.40 & 3.30 & 40.88 & 12.02 & 70.83 & 45.59 \\
     RAG Agent ($\mathcal{D}_{\text{rag}}$) & 41.75 & 11.00 & 11.00 & 45.01 & 52.00 & 42.70 & 55.86 & 55.50 & 66.05 & 54.40 & 64.32 & 56.06 & 5.94 & 75.84 & 37.39 \\
    \arrayrulecolor{black}\midrule
    \multicolumn{16}{l}{\textit{\textbf{Data-level Consolidation}}} \\
    \arrayrulecolor{black!20}\midrule
     Data Mixing ($\mathcal{D}_{\text{all}}$) & \cellcolor{top1bg}56.31 & \cellcolor{top1bg}16.00 & 10.00 & \cellcolor{top2bg}62.11 & 71.69 & \cellcolor{top1bg}43.50 & \cellcolor{top1bg}54.87 & \cellcolor{top1bg}54.40 & \cellcolor{top1bg}64.16 & \cellcolor{top1bg}50.40 & \cellcolor{top1bg}61.22 & \cellcolor{top1bg}71.92 & 0.99 & 73.96 & 36.17 \\
    \arrayrulecolor{black}\midrule
    \multicolumn{16}{l}{\textit{\textbf{Parameter-level Consolidation}}} \\
    \arrayrulecolor{black!20}\midrule
     \cellcolor{bgblue}Average & \cellcolor{top2bg}55.34 & \cellcolor{top3bg}13.00 & \cellcolor{top1bg}21.00 & 55.91 & 68.69 & 14.80 & 28.76 & 20.80 & 38.70 & 15.20 & 32.62 & 60.02 & 1.37 & 76.88 & 44.07 \\
     \cellcolor{bgblue}SLERP & 49.51 & 10.00 & 16.00 & 58.78 & 71.96 & 11.40 & 24.38 & 17.70 & 36.42 & 12.80 & 30.78 & 55.66 & 1.22 & 74.80 & 44.68 \\
     \cellcolor{bgblue}MetaGPT & 46.60 & \cellcolor{top3bg}13.00 & 13.00 & 58.13 & 70.05 & 16.60 & 26.12 & 23.70 & 35.31 & 20.00 & 29.81 & 57.13 & 0.44 & 72.30 & 40.73 \\
     \cellcolor{bgblue}LiNeS & 46.60 & 11.00 & 16.00 & 53.33 & 74.29 & 24.60 & 39.10 & 30.70 & 47.85 & 27.20 & 44.71 & 62.60 & \cellcolor{top2bg}0.09 & 68.55 & 39.21 \\
     \cellcolor{bggreen}DARE & 52.43 & 9.00 & 14.00 & \cellcolor{top1bg}62.29 & 71.51 & 15.10 & 29.20 & 21.80 & 40.54 & 27.20 & 43.54 & 59.91 & 0.26 & 74.59 & 39.51 \\
     \cellcolor{bggreen}Breadcrumbs & 19.42 & 1.50 & 3.00 & 45.66 & 62.34 & 14.80 & 22.39 & 15.40 & 23.57 & 20.80 & 30.20 & 33.17 & 7.75 & 75.64 & 47.42 \\
     \cellcolor{bggreen}TIES & 38.83 & \cellcolor{top2bg}13.50 & \cellcolor{top2bg}20.00 & 59.52 & 70.24 & 11.10 & 29.83 & 16.10 & 36.65 & 17.60 & 37.02 & 58.67 & 0.90 & 70.64 & 40.12 \\
     \cellcolor{bggreen}Consensus TA & 8.74 & 1.50 & 1.00 & 31.79 & 30.04 & 6.50 & 13.84 & 6.50 & 10.05 & 7.20 & 16.28 & 18.82 & 4.05 & 23.74 & 8.51 \\
     \cellcolor{bggreen}TADrop & 45.63 & 10.50 & 16.00 & 56.28 & 70.60 & 16.10 & 34.02 & 19.40 & 41.52 & 14.40 & 36.32 & 57.46 & \cellcolor{top3bg}0.23 & 73.33 & 41.95 \\
     \cellcolor{bggreen}CABS & 44.66 & 10.00 & 14.00 & 56.19 & 71.23 & 0.50 & 0.85 & 0.50 & 0.00 & 0.00 & 0.62 & 38.59 & 13.40 & 67.50 & 33.74 \\
     \cellcolor{bggreen}PCB Merging & 50.49 & 12.00 & 16.00 & 60.07 & \cellcolor{top3bg}74.32 & 18.80 & 28.50 & 22.80 & 34.81 & 20.00 & 30.78 & 60.36 & 0.64 & 78.96 & 45.59 \\
     \cellcolor{bggreen}DELLA & 6.80 & 1.00 & 2.00 & 30.04 & 29.58 & 6.40 & 13.03 & 6.10 & 13.81 & 8.00 & 17.19 & 18.45 & 4.82 & 34.38 & 11.25 \\
     \cellcolor{bggreen}SCE & 27.18 & 3.00 & 18.00 & 52.49 & 55.93 & 25.90 & 35.17 & 37.40 & 49.72 & 28.80 & 37.96 & 50.79 & 2.02 & \cellcolor{top3bg}79.58 & 46.81 \\
     \cellcolor{bggreen}WIDEN & 17.48 & 2.50 & 2.00 & 34.56 & 41.47 & 11.90 & 21.41 & 10.30 & 20.57 & 12.80 & 24.67 & 25.26 & 4.42 & 42.92 & 16.41 \\
     \cellcolor{bggreen}RAM+ & 50.49 & 12.00 & 17.00 & 56.93 & 71.05 & 10.90 & 28.80 & 19.10 & 39.25 & 12.80 & 32.30 & 57.56 & 1.04 & 73.74 & 41.33 \\
     \cellcolor{bggreen}TSV & 47.57 & \cellcolor{top3bg}13.00 & \cellcolor{top1bg}21.00 & 53.88 & 51.91 & 13.50 & 27.74 & 17.70 & 33.95 & 20.80 & 34.83 & 54.21 & 1.81 & 75.23 & 41.64 \\
     \cellcolor{bggreen}ISO-CTS & 0.97 & 0.00 & 0.00 & 18.67 & 9.00 & 0.00 & 0.00 & 0.00 & 0.00 & 0.00 & 0.00 & 10.73 & 5.58 & 0.00 & 0.00 \\
     \cellcolor{bggreen}IMPART & \cellcolor{top3bg}53.39 & 10.50 & 15.00 & 58.78 & 65.06 & 10.30 & 26.25 & 18.80 & 37.02 & 17.60 & 36.46 & 56.03 & 0.63 & 70.41 & 39.01 \\
     \cellcolor{bggreen}WUDI & 52.43 & 12.00 & \cellcolor{top3bg}19.00 & 58.96 & 71.05 & 20.70 & 32.56 & 30.00 & 43.15 & 20.80 & 34.19 & 63.32 & 0.26 & 74.37 & 40.12 \\
     \cellcolor{bggreen}DC-Merge & 49.51 & 8.00 & 13.00 & 58.50 & 70.69 & \cellcolor{top3bg}35.10 & \cellcolor{top2bg}46.45 & \cellcolor{top3bg}41.20 & \cellcolor{top3bg}52.68 & \cellcolor{top2bg}45.60 & \cellcolor{top2bg}54.71 & 64.26 & 1.38 & \cellcolor{top1bg}80.62 & \cellcolor{top3bg}48.33 \\
     \cellcolor{bggreen}OrthoMerge-C & 47.57 & 12.50 & 12.00 & 59.06 & 70.87 & 30.30 & 42.83 & 38.50 & 51.58 & \cellcolor{top3bg}40.80 & \cellcolor{top3bg}54.43 & \cellcolor{top3bg}65.60 & 0.63 & 78.95 & 47.72 \\
     \cellcolor{bggreen}OrthoMerge-G & \cellcolor{top2bg}55.34 & 12.00 & 15.00 & 59.33 & 71.60 & 16.40 & 29.12 & 19.90 & 38.92 & 21.60 & 35.81 & 60.34 & 0.48 & 75.00 & \cellcolor{top1bg}49.54 \\
     \cellcolor{bgred}AdaMerging & 6.80 & 0.00 & 7.00 & 0.00 & 9.17 & 0.00 & 0.00 & 0.00 & 0.00 & 0.00 & 0.00 & 10.46 & 16.72 & 0.00 & 11.25 \\
     \cellcolor{bgred}RegMean++ & 48.54 & 12.50 & 16.00 & 58.96 & \cellcolor{top2bg}76.31 & \cellcolor{top2bg}35.50 & \cellcolor{top3bg}46.12 & \cellcolor{top2bg}49.10 & \cellcolor{top2bg}59.40 & \cellcolor{top3bg}40.80 & 50.76 & \cellcolor{top2bg}70.21 & 0.49 & 74.16 & 45.59 \\
     \cellcolor{bgred}CAT Merging & 51.46 & 12.50 & 15.00 & \cellcolor{top3bg}61.83 & 71.87 & 24.60 & 34.58 & 37.00 & 47.65 & 25.60 & 38.12 & 65.12 & \cellcolor{top1bg}0.01 & \cellcolor{top2bg}80.01 & \cellcolor{top2bg}48.94 \\
     \cellcolor{bgred}SLTA & 47.57 & 12.50 & 13.00 & 54.16 & \cellcolor{top1bg}76.68 & 11.00 & 23.88 & 16.30 & 34.27 & 16.00 & 32.15 & 55.09 & 1.09 & 74.79 & 45.28 \\
    \arrayrulecolor{black}\bottomrule
    \end{tabular}
    }
\end{table*}

\begin{table}[tbp]
\centering
\small
\caption{Definitions of the 11 information-seeking behavior categories used in Figure~\ref{fig:behavior}.}
\setlength{\tabcolsep}{6pt}
\begin{tabular}{p{0.06\linewidth} p{0.22\linewidth} p{0.66\linewidth}}
\hline
\textbf{ID} & \textbf{Category} & \textbf{Definition} \\
\hline
A & Q-Decomp & \textbf{Query decomposition}: breaking a complex information need into multiple sub-questions or sub-goals that can be solved sequentially. \\
B & Q-Refine & \textbf{Query refinement}: modifying or reformulating the query based on intermediate results (e.g., adding constraints, using synonyms, changing phrasing) to improve retrieval relevance. \\
C & Hypoth & \textbf{Hypothesis generation}: forming a tentative answer or explanation before full verification, typically as a working theory to guide further search. \\
D & EvidenceV & \textbf{Evidence verification}: checking whether retrieved information supports (or contradicts) the current hypothesis/answer, including consistency checks against the original question. \\
E & CrossRef & \textbf{Cross-reference validation}: comparing or triangulating information across different sources, sections, or passages to confirm correctness or resolve conflicts. \\
F & GapID & \textbf{Gap identification}: explicitly recognizing missing, unclear, or insufficient information needed to answer the question, and stating what is still required. \\
G & ErrRecov & \textbf{Error recovery}: handling failed, irrelevant, or noisy retrieval/search attempts by applying corrective actions (e.g., retrying with new terms, switching strategy). \\
H & Backtrck & \textbf{Backtracking}: abandoning an unproductive line of search and returning to an earlier step, alternative hypothesis, or previous plan to continue progress. \\
I & InfoSynth & \textbf{Information synthesis}: integrating evidence from multiple retrieved pieces into a coherent understanding, combining complementary facts rather than relying on a single snippet. \\
J & AnsForm & \textbf{Answer formulation}: organizing and presenting the final response clearly (e.g., structured reasoning, ordered steps, concise conclusion) based on accumulated evidence. \\
K & Constr & \textbf{Constraint satisfaction}: ensuring the final answer satisfies explicit requirements and constraints (format, scope, conditions, units, time range, etc.) stated by the user/task. \\
\hline
\end{tabular}

\label{tab:behavior_categories}
\end{table}

\section{Single-Capability Champions Across Scenarios}
\label{app:single_capability}

Figure~\ref{fig:single_capability} visualises per-capability rankings of all merging methods across the three training scenarios. We sort methods within each capability panel by their \emph{mean rank} across scenarios; cell colour encodes the rank in each individual scenario (greener $=$ better). Three observations stand out: \textbf{(i)} the Web capability champion shifts substantially across scenarios---\textit{Average} ascends to \#1 in the 4B and LoRA settings while TIES/CABS lead on 30B-A3B; \textbf{(ii)} Doc is dominated by data-dependent methods, with \textit{RegMean++} ranking first or near-first in every scenario; \textbf{(iii)} RAG is essentially monopolised by the \textit{RAG Agent} expert, the only entity that ranks \#1 on RAG in all three scenarios. These patterns complement the aggregated views in Tables~\ref{tab:performance_comparison}, \ref{tab:performance_comparison_30b_lora}, and \ref{tab:performance_comparison_4b}, and motivate the cross-scenario stability analysis in \S\ref{sec:empirical} (Figure~\ref{fig:stability}).

\section{Four-Class Method Taxonomy via $G \times \mathrm{Imb}$}
\label{app:four_class}

Jointly inspecting the Composite Score $G$ and the Imbalance Score $\mathrm{Imb}$ decomposes the evaluated methods into four engineering classes with distinct remediation strategies (using the 30B-A3B numbers from Table~\ref{tab:performance_comparison} as the reference scenario; the same partitions hold qualitatively on 4B and LoRA, see Tables~\ref{tab:performance_comparison_4b} and~\ref{tab:performance_comparison_30b_lora}):

\begin{description}
\item[All-rounders ($G$ high, $\mathrm{Imb}$ low).] Methods that balance the three task domains. Representatives: Data Mixing ($G=71.06$, $\mathrm{Imb}=0.37$), RegMean++ ($G=67.46$, $\mathrm{Imb}=0.91$), OrthoMerge-C ($G=58.43$, $\mathrm{Imb}=0.02$), CAT Merging ($G=57.51$, $\mathrm{Imb}=0.12$), and OrthoMerge-G ($G=56.07$, $\mathrm{Imb}=0.19$). \emph{Engineering action:} directly deployable.
\item[Specialists ($G$ moderate, $\mathrm{Imb}$ high).] Methods that excel on a single domain at the cost of the others. Representatives: RAG Agent ($G=30.77$, $\mathrm{Imb}=25.32$, RAG Score $\approx 87$ vs.\ Web Score $\approx 20$); also Web Agent and Doc Agent on the 4B scenario after RAG erasure. \emph{Engineering action:} retain for the dominant capability; pair with complementary components for missing ones.
\item[Collapsed ($G$ low, $\mathrm{Imb}$ low).] Methods that fail uniformly across domains. Representatives: ISO-CTS ($G=4.77$, all three task scores below $10$) and Consensus TA ($G=14.64$, all three task scores below $25$). \emph{Engineering action:} structural failure; usually unrecoverable without redesigning the merging operator.
\item[Imbalanced underperformers ($G$ low, $\mathrm{Imb}$ high).] Methods that fail on one or two domains while remaining acceptable on another. Representatives: SCE on 30B-A3B ($G=47.04$, $\mathrm{Imb}=9.75$, Web Score $\approx 27$) and AdaMerging on 4B ($G=10.46$, with collapsed Doc and Web). \emph{Engineering action:} targeted intervention---fix the missing capability via auxiliary data or a complementary expert.
\end{description}

The cross-scenario migrations of class memberships further reveal robustness profiles: DC-Merge moves from \textit{Collapsed} at 30B to \textit{All-rounder} at 4B and LoRA; WUDI moves from \textit{All-rounder} at 30B to \textit{Imbalanced} (in fact almost \textit{Collapsed}) under LoRA; and AdaMerging is \textit{Imbalanced} at 30B but \textit{Collapsed} at 4B. These migrations align with the cross-scenario failure analysis in \S\ref{sec:failure}.

\section{Potential Risks}
A stronger and more general information-seeking agent would be constructed, which inherits and may amplify the risks of their per-domain components, or synthesize some unexpected behavior. Unified web, document, and knowledge-base capabilities lower the cost of large-scale information gathering and could be misused for surveillance, targeted phishing, or automated extraction of private content. They may also compound hallucinations across heterogeneous sources when retrieval grounding is weak.

\end{document}